\documentclass{article}

\PassOptionsToPackage{numbers,sort&compress}{natbib}
\PassOptionsToPackage{preprint}{neurips_2026}
\usepackage{neurips_2026}

\usepackage[utf8]{inputenc} 
\usepackage[T1]{fontenc}    
\usepackage{hyperref}       
\usepackage{url}            
\usepackage{booktabs}       
\usepackage{amsfonts}       
\usepackage{nicefrac}       
\usepackage{microtype}      
\usepackage{xcolor}         
\usepackage{graphicx}
 \usepackage{amsmath}
 \usepackage{subcaption}
\usepackage{enumitem}
\usepackage{float}

\title{Molecules Meet Language: Confound-Aware Representation Learning and Chemical Property Steering in Transformer-VAE Latent Spaces}

\author{%
  Zakaria Elabid\\
  Helmholtz-Zentrum Dresden-Rossendorf\\
  Center for Advanced Systems Understanding\\
  \texttt{z.elabid@hzdr.de}\\
  \And
  Jan Andrzejewski\\
  Helmholtz-Zentrum Dresden-Rossendorf\\
  Center for Advanced Systems Understanding\\
  j.andrzejewski@hzdr.de
  \And 
  Bartosz Brzoza\\
  Helmholtz-Zentrum Dresden-Rossendorf\\
  Center for Advanced Systems Understanding\\
  b.brzoza@hzdr.de
  \And 
  Attila Cangi\\
  Helmholtz-Zentrum Dresden-Rossendorf\\
  Center for Advanced Systems Understanding\\
  \texttt{a.cangi@hzdr.de}
}

\begin{document}

\maketitle

\begin{abstract}
Molecular generative models often assume meaningful latent geometry, but apparent property predictability can reflect sequence-level shortcuts rather than chemical organization. We study this issue in an unsupervised autoregressive Transformer-VAE trained on SELFIES. After training, we freeze the model, fit linear probes to RDKit descriptors, and use the probe weights as candidate global steering directions. To separate chemical signal from SELFIES artifacts, we introduce a confound-aware evaluation based on residualization, confound-direction alignment analysis, and decoded-molecule traversal. This is necessary because SELFIES length, branch tokens, ring tokens, and token entropy are strongly encoded in the latent space. Under this confound-aware evaluation, we find robust monotonic steering for cLogP, FractionCSP3, HeavyAtomCount, TPSA, BertzCT, and HBA. Nonlinear probes further show that some properties admit stable global directions, while others are better described by local latent gradients. Overall, our results show that chemically meaningful steering can emerge in entangled molecular latent spaces, but only when validated through decoded molecules and controlled for representation-level confounds.
\end{abstract}


\section{Introduction}

Molecular design is a central problem in drug discovery and materials science, where the objective is to generate or optimize chemical structures with desired properties \cite{sanchezlengeling2018inverse,vamathevan2019applications,du2022molgensurvey}. Although molecules are naturally represented as labeled graphs, molecular line notations make them accessible to sequence models. SMILES provides a compact chemical language for molecular graphs \cite{weininger1988smiles}, and its non-unique forms have been used as data augmentation in molecular learning \cite{bjerrum2017smiles}. SELFIES addresses a different but crucial issue for generation: arbitrary SELFIES strings decode to valid molecules under the representation grammar \cite{krenn2020selfies}. In this work, molecular language is therefore used as the training interface, while chemical validity and molecular properties are evaluated after decoding into graph-based molecular objects.

Language modeling has become a general mechanism for learning representations from unlabeled sequences. Transformers enabled scalable sequence modeling through attention \cite{vaswani2017attention}, while large-scale pretraining demonstrated that useful representations can emerge without task-specific supervision \cite{devlin2019bert}. Related principles have been observed in biological sequences \cite{rives2021biological} and chemical strings, including SMILES Transformer fingerprints \cite{honda2019smiles}, ChemBERTa-style molecular pretraining \cite{chithrananda2020chemberta}, SELFIES-based Transformer representations \cite{yuksel2023selformer}, and autoregressive molecular generation with MolGPT \cite{bagal2021molgpt}. These works support the view that molecular strings contain learnable chemical regularities. However, they do not fully answer whether an unsupervised molecular language model learns a continuous latent space with chemically meaningful and steerable directions.

Molecular generation and optimization have followed complementary directions. Optimization searches in discrete molecular graph or string space using reinforcement learning, genetic search, or graph-based policies \cite{you2018gcpn,zhou2019optimization,jensen2019graph,fu2022reinforced}. Generative approaches instead learn a distribution over molecules and use the learned representation for sampling, interpolation, or optimization, including molecular VAEs, junction-tree models, normalizing flows, and graph diffusion models \cite{gomezbombarelli2018automatic,jin2018junction,zang2020moflow,vignac2023digress}. Latent-space optimization exploits trained generators as continuous search spaces, for example through Bayesian optimization, latent inceptionism, linear molecular manipulation, or learned latent flows \cite{griffiths2020constrained,eckmann2022limo,du2023chemspace,wei2024chemflow}. These methods show that latent molecular spaces can support optimization, in a nontrivial manner: a useful generator does not necessarily provide globally meaningful chemical directions.

This paper studies that question in a deliberately constrained setting. We train a slot-based autoregressive Transformer-VAE on SELFIES sequences using only reconstruction and latent regularization. After training, the encoder and decoder are frozen. We then fit linear probes to RDKit-computed molecular descriptors and use the learned probe weights as candidate global steering directions. Thus, property labels are not used to train the generative model; they are used only post hoc to interrogate the latent space. The main evaluation is not whether a descriptor can be predicted from a molecular string, but whether moving through the frozen latent space along a probe-derived direction changes decoded molecules in the intended chemical direction.

A central risk is that a molecular language model may encode sequence-level shortcuts rather than chemical structure. SELFIES improves validity, but it remains a token sequence whose length, branch tokens, ring tokens, and token entropy can correlate with molecular size, topology, and descriptor values. We therefore evaluate latent directions under a confound-aware protocol: descriptors must be predictable from the frozen latent representation, robust to measured SELFIES-level confounds, and validated after decoding through molecular properties rather than latent probe scores alone.

\begin{figure}[t]
    \centering
    \includegraphics[width=0.8\textwidth]{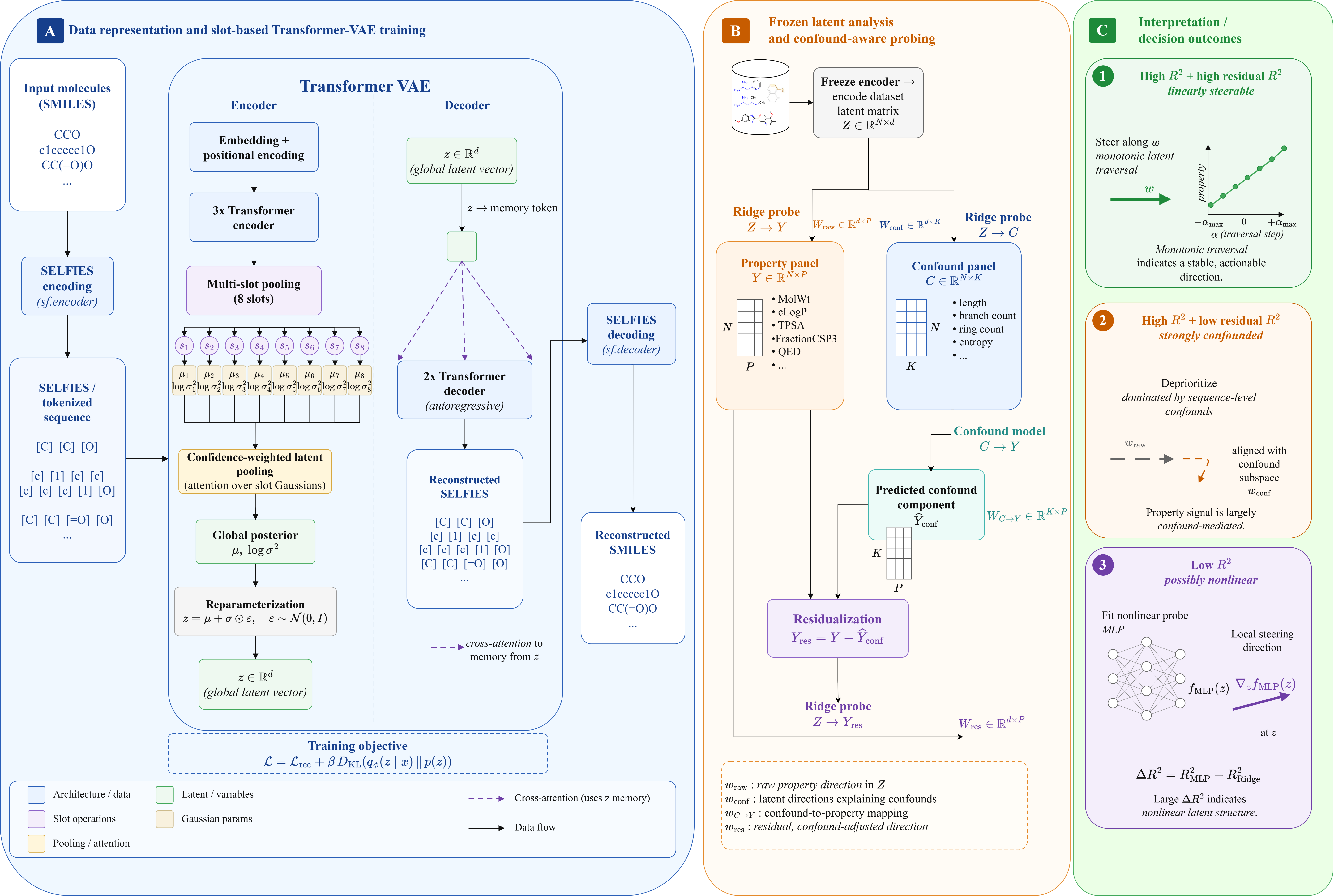}
    \caption{
    Overview of the proposed framework. 
    \textbf{A.} SMILES are converted to SELFIES, tokenized, and used to train an autoregressive Transformer-VAE. 
    \textbf{B.} The encoder is frozen and the latent space is probed for molecular properties $Y$ and SELFIES-level confounds $C$; residualization removes the component predictable from the confounds. 
    \textbf{C.} Raw and residual $R^2$ values, together with decoded-molecule traversals, separate linearly steerable properties from confounded or nonlinear ones.
    }
    \label{fig:workflow}
\end{figure}

The full framework is summarized in Figure \ref{fig:workflow}. Our contributions are as follows:
\begin{itemize}[leftmargin=1.5em,itemsep=0.1em,topsep=0.1em,parsep=0pt,partopsep=0pt]
    \item We frame molecular property steering as post hoc interpretation of an unsupervised molecular language model trained on SELFIES.

    \item We connect linear probing to latent chemical control by treating probe weights as global steering axes and validating them through decoded molecules.

    \item We introduce a confound-aware evaluation that separates chemical organization from SELFIES-level shortcuts such as token length, branch-token count, ring-token count, and token entropy.

    \item We show that useful global directions can emerge in an entangled latent space, with cLogP, FractionCSP3, BertzCT, TPSA, and HBA exhibiting robust monotonic steering. 

    \item We use nonlinear probes diagnostically to distinguish globally linear steerability from local or nonlinear latent organization.
\end{itemize}

\section{Background}

\subsection{Molecular representations and chemical language}

Molecules are graphs whose atoms and bonds define chemical structure. Graph-based generators model this structure directly, avoiding arbitrary traversal order, and have been developed through junction-tree decomposition, implicit graph generation, invertible graph flows, and discrete graph diffusion \cite{jin2018junction,cao2018molgan,zang2020moflow,vignac2023digress}. Their advantage is structural explicitness; their difficulty is that valid molecular generation remains a constrained discrete problem involving rings, graph size, and graph isomorphism.

String representations expose the same chemical objects to sequence models. SMILES is compact, expressive, and widely supported, but valid generation requires learning a constrained formal language involving branches, ring closures, aromaticity, and stereochemical conventions \cite{weininger1988smiles}. Canonicalization can impose a deterministic form, while non-canonical SMILES can also be exploited as augmentation \cite{bjerrum2017smiles}. SELFIES changes this trade-off by embedding stronger validity constraints into the representation grammar, making it particularly suitable for sequence-based generative modeling~\cite{krenn2020selfies}. 

\subsection{Chemical language models and latent molecular generation}

Transformers provide a natural architecture for molecular strings because autoregressive decoding can model token sequences one symbol at a time \cite{vaswani2017attention}. More broadly, self-supervised language modeling has shown that useful structure can be recovered from unlabeled sequences in natural language and biology \cite{devlin2019bert,rives2021biological}. Chemical language models extend this principle to molecular strings through unsupervised SMILES pretraining, large-scale molecular Transformers, SELFIES-based representation learning, and GPT-style molecular generation \cite{honda2019smiles,chithrananda2020chemberta,yuksel2023selformer,bagal2021molgpt}.

Latent-variable molecular models add a continuous representation between discrete molecules and decoded structures. The VAE framework provides an encoder--decoder formulation for such continuous latent spaces \cite{kingma2013autoencoding}, and molecular VAEs showed that this geometry can support interpolation and optimization in chemical design \cite{gomezbombarelli2018automatic}. However, reconstruction quality or latent smoothness does not imply semantic disentanglement. A model may organize latent space around dataset frequency, sequence length, or molecular size rather than chemically interpretable factors. Our work therefore asks whether an unsupervised Transformer-VAE latent space contains simple directions that remain meaningful after confound-aware analysis and decoded-molecule validation.

\subsection{Latent traversal and molecular optimization}

Latent optimization methods treat a trained molecular generator as a continuous search space. Constrained Bayesian optimization and latent inceptionism use surrogate or oracle-driven objectives to search VAE latent spaces for improved molecules \cite{griffiths2020constrained,eckmann2022limo}. Linear manipulation methods impose a stronger interpretability assumption: a reusable direction in latent space should correspond to a molecular change across many seeds. For example, linear decision boundaries have been used to define molecular manipulation directions through their normal vectors \cite{du2023chemspace}. ChemFlow generalizes latent traversal by learning vector fields that transport molecular latent distributions toward property or diversity objectives \cite{wei2024chemflow}.

Our objective is simpler and more diagnostic. We do not learn a flow or optimize a supervised property objective during generator training. Instead, we test whether linear probes fitted after unsupervised training reveal globally reusable directions in the frozen latent space. This makes the method efficient and interpretable, but also imposes a stricter validation burden: a direction must produce meaningful decoded molecular changes, not merely high latent-space predictor scores.

\subsection{Probing representations and controlling shortcuts}

Probing is a standard method for analyzing what information is accessible in learned representations. Linear probes were introduced as a way to test linearly available information in neural features \cite{alain2016understanding}, but probe accuracy alone can be misleading without capacity controls and appropriate baselines \cite{hewitt2019designing,belinkov2022probing}. This distinction is central for molecular steering. A linear probe is not used here because it is the strongest predictor, but because its weight vector defines a single global direction. Nonlinear probes can reveal additional recoverable information, but their gradients are local and model-dependent, making them better suited for diagnosing nonlinearity than for claiming global steerability.


\section{Methodology}

\subsection{Dataset Representation and Property Panel}
\label{sec:dataset_properties}

Molecules are represented as SELFIES token sequences derived from the original SMILES strings. Raw SMILES provide compact molecular encodings, but their sequence form depends on traversal conventions and can introduce syntactic variability that is not necessarily chemically meaningful. We therefore use SELFIES as the training representation, since they provide a grammar-constrained molecular language while remaining directly tokenizable for sequence modeling \citep{krenn2020selfies}.

Each molecule is encoded as
$
x = (x_1,\ldots,x_T),
$
where $x_t$ denotes a SELFIES token. Padding is used only for batching, where $T$ is the maximum length after padding. The Transformer-VAE is trained on these tokenized SELFIES sequences, while all downstream molecular descriptors are computed from the corresponding RDKit molecular graph.

For latent-space analysis, we construct a property matrix
$
Y \in \mathbb{R}^{N \times P},
$
where each row corresponds to a molecule, totaling N molecules, and each column corresponds to an RDKit-computed scalar descriptor, totaling P descriptors. These descriptors cover size, polarity, flexibility, ring topology, lipophilicity, three-dimensionality, topological complexity, synthetic accessibility, and drug-likeness. The full descriptor list, definitions, RDKit implementations, and tier assignments are provided in Appendix~\ref{app:property_panel}. This separation keeps the main analysis focused on latent predictability and steerability, while the appendix documents the chemical interpretation of each property.
\subsection{Transformer-VAE Formulation}

Given a SELFIES-tokenized molecule \(x=(x_1,\ldots,x_T)\), the model encodes the token sequence with a Transformer encoder,
$
H = \mathrm{Enc}_{\theta}(x) \in \mathbb{R}^{T \times h}
$, where $\theta$ denotes the weights and $h$ denotes the hidden size. The self-attention mechanism allows the representation to capture dependencies between distant SELFIES tokens, such as branches, ring closures, and functional substructures.

To obtain a molecular latent representation, the encoded token states are aggregated using learned multi-slot pooling. The slot representations are mapped to Gaussian posterior components and combined into a single global posterior,
$
q_{\theta}(z \mid x)
=
\mathcal{N}
\left(
\mu,
\operatorname{diag}(\sigma^2)
\right).
$
Here, \(\mu \in \mathbb{R}^d\) and \(\sigma \in \mathbb{R}^d_{>0}\) denote the mean and standard deviation of the posterior over the \(d\)-dimensional latent variable \(z\).
Details of the slot construction and confidence-weighted aggregation are provided in Appendix~\ref{app:architecture}.

The latent variable is sampled using the reparameterization trick,
$
z
=
\mu
+
\sigma \odot \epsilon,
\epsilon \sim \mathcal{N}(0,I).
$

A Transformer decoder reconstructs the SELFIES sequence autoregressively from the latent vector.
During training, the decoder receives the shifted ground-truth sequence under teacher forcing, while causal masking prevents access to future tokens. Cross-attention to the latent memory routes reconstruction through \(z\), which is later used for property probing, confound analysis and traversal.

The model is trained with the VAE objective: 
$
\mathcal{L}
=
\mathcal{L}_{\mathrm{rec}}
+
\beta
D_{\mathrm{KL}}
\left(
q_{\theta}(z \mid x)
\;\middle\|\;
\mathcal{N}(0,I)
\right),
$

where \(\mathcal{L}_{\mathrm{rec}}\) is the reconstruction loss, and $\beta$ is the regularization coefficient. The reconstruction term preserves molecular identity, while the KL term regularizes the posterior toward a smooth latent prior.

\subsection{VAE Benchmarking Metrics}

We evaluate the VAE with five generative checks: reconstruction, novelty, validity, uniqueness, and family retention. Reconstruction measures exact sequence recovery, while novelty measures the fraction of generated molecules absent from the training set, providing a basic check against memorization. Validity is computed by decoding sampled latent vectors to SELFIES, converting them to SMILES, and applying RDKit sanitization. Uniqueness is computed over the valid canonicalized SMILES, so duplicate molecular structures are not counted separately.

Family retention measures whether interpolation paths between molecules from the same functional family preserve that family across decoded steps. Full details are provided in Appendix~\ref{app:interpolation_smoothness}.


\subsection{Confound-Aware Latent Property Analysis}
\subsubsection{Property and Confound Prediction from Latent Space}

For each molecule \(x_i\), the encoder produces a latent representation
$
z_i = E(x_i) \in \mathbb{R}^d.
$
Each latent point is associated with two sets of scalar targets: a property panel
$
Y \in \mathbb{R}^{N \times K},
$
containing $K$ molecular descriptors, and a confound panel
$
C \in \mathbb{R}^{N \times M},
$
containing $M$ sequence-level statistics derived from the SELFIES representation.

We include the confound panel because molecular properties can correlate with superficial string-level features, such as sequence length or special-token counts. A probe trained directly on \(z_i\) may therefore achieve high predictive performance by exploiting these shortcut variables rather than chemically meaningful latent organization.

To quantify this, we fit one linear probe per target using frozen latent representations:
$\hat y_i^{(k)} = w_k^\top z_i + b_k$,
$\hat c_i^{(m)} = u_m^\top z_i + a_m
$. Here, $y$ and $c$ denote property and confound targets, $w_k$ and $u_m$ are probe weights, and $b_k$ and $a_m$ are intercepts.
Performance is evaluated on held-out data using \(R^2\).

This step is diagnostic. High \(R^2\) for a molecular property indicates that the latent space contains predictive information, but not necessarily that the signal is chemically specific. If the same latent space also predicts confounds that are correlated with the property, the apparent property signal may be shortcut-driven. This motivates the residualization procedure introduced next, where the component of each property explainable from the confound panel is removed before interpreting latent-property directions.

\subsubsection{Residualization against SELFIES Confounds}

Let \(Z\), \(Y\), and \(C\) denote the frozen latent representations, molecular property panel, and SELFIES confound panel. For each property \(y\), we fit a confound-only predictor
$
\hat y^{(C)} = g(C),
$
and define the residualized target:
$
y_{\mathrm{res}} = y - \hat y^{(C)}.
$

Latent probes are then evaluated on \(y_{\mathrm{res}}\) rather than on the raw property:
$
\widehat{y}_{\mathrm{res}}
=
w_{\mathrm{res}}^\top z + b_{\mathrm{res}}.
$

This tests whether latent predictability remains after removing the component explainable from SELFIES-level artifacts. A large performance drop indicates confound-mediated signal, while stable performance suggests that the latent space captures information beyond the measured shortcuts. Residualized targets are therefore used only diagnostically, to shortlist properties for traversal.

\subsection{Linear Probes as Global Property Directions}

We use linear probes to test whether a property is organized along a steerable latent direction. Given frozen latent codes
$
z_i = E(x_i) \in \mathbb{R}^d
$, where $d$ is the model dimension,
and scalar property values \(y_i\), we fit a linear probe
$
\hat y_i = w^\top z_i + b.
$
The probe weight \(w\) is then interpreted as the global latent direction associated with the property.

For a perturbation
$
z' = z + \epsilon w,
$
the predicted property changes as
$
\hat y(z') 
= \hat y(z) + \epsilon \|w\|_2^2.
$
Thus, moving along \(w\) increases the predicted property under the linear approximation, while moving along \(-w\) decreases it. Since
$
\nabla_z \hat y(z) = w,
$
the probe defines a single global direction field.

Successful traversal along \(w\) therefore indicates that the property is approximately globally linear in latent space. Failure or instability suggests that the property is weakly encoded, local, or not organized along a single global direction. Additionally, once a property direction has been learned, improving the surrogate objective does not
require iterative gradient-based optimization, black-box search, or repeated property-oracle
queries in latent space. 
Further discussion is given in Appendix~\ref{app:monotonic}.

\subsection{Nonlinear Probe Formulation and Local Steering}

The linear probe defines a single global direction only when a property is approximately affine in the latent space. Properties with weak or moderate linear performance may still be encoded in the latent representation, but arranged along curved or locally varying manifolds. We therefore use nonlinear probes as a diagnostic extension of the linear direction study, focusing on the remaining non-trivial descriptors after excluding strongly size-confounded targets. 

For each selected property, we fit a multilayer perceptron on the same frozen latent representations and train/validation/test splits used for the linear probes. Latent vectors are standardized using training-set statistics,
$
\tilde z_i
=
\frac{z_i-\mu_Z}{\sigma_Z},
$
and targets are standardized analogously during optimization. 

For a scalar target $y$, the nonlinear probe is
$
\hat y_i
=
f_{\phi}(\tilde z_i),
$
where $f_{\phi}$ is a multi-layer perceptron (MLP),
with ReLU nonlinearities. The model is trained by minimizing mean-squared error on the training set,
with model selection by validation $R^2$ and final reporting on the held-out test set.

The same procedure is applied to both raw and residual targets. For the raw target,
$
\hat y_i = f_{\phi,y}(\tilde z_i),
$
whereas for the confound-corrected target,
$
\hat y_{\mathrm{res},i}
=
f_{\phi,y_{\mathrm{res}}}(\tilde z_i).
$
Thus, the raw MLP measures whether additional nonlinear property information is present in the latent code, while the residual MLP tests whether that nonlinear information remains after removing SELFIES-level confounds.

Additional information about the use of MLP nonlinear probing for gradient optimization of properties from the latent can be found in the Appendix \ref{app:nonlinear gradient}.

\section{Experiments}

\subsection{Experimental Setup}

We train and evaluate Transformer-VAE models on 794,403 RDKit valid SELFIES molecules. SELFIES strings are tokenized with start and end tokens, and padded to a maximum sequence length of 77. The saved tokenizer contains 110 tokens, including PAD, SOS, and EOS; the autoregressive implementation additionally reserves a MASK token. 
The dataset is split into train, validation, and test partitions using an 80/10/10 split.

The molecular descriptor panel contains the properties described in Appendix \ref{app:property_panel}. To control for representation-level artifacts, we also compute a SELFIES confound panel consisting of token length, branch-token count, ring-token count, and token entropy.


\subsection{Model Variants and Ablation Settings}

We compare three Transformer-VAE variants. The two non-autoregressive models serve as architectural controls, while the autoregressive model is used as the main model for the confound-aware property analysis. Table \ref{tab:model_variants} summarizes the details of the architectures used in this ablation study.

\begin{table}[t]
\centering
\caption{Transformer-VAE model variants.}
\label{tab:model_variants}
\resizebox{\textwidth}{!}{
\begin{tabular}{lccccccc}
\toprule
Model name & Decoder type & Latent dim. & Hidden size & Encoder blocks & Decoder blocks & Slot pooling & Decoder attention \\
\midrule
Linear Attention Non-AR & Non-AR & 512 & 256 & 1 & 1 & Linear slot pooling & Linear cross-attention \\
Simple Attention Non-AR & Non-AR & 256 & 256 & 1 & 1 & Multi-head slot pooling & Standard cross-attention \\
Autoregressive MultiSlotting  & AR & 256 & 256 & 3 & 2 & Multi-slot pooling & Causal Transformer decoder with latent memory \\
\bottomrule
\end{tabular}
}
\end{table}

The comparison is designed to separate reconstruction quality from latent-space utility. High reconstruction is necessary for a useful molecular VAE, but it does not by itself show that molecular properties are organized as chemically meaningful latent directions.

\subsection{Model Selection and Baseline Latent Quality}

Table~\ref{tab:recon_gen_interp} summarizes reconstruction for test/val, and generation and interpolation results, defined previously in the methodology for N=5000 randomly sampled latents. Linear Attention Non-AR achieves the strongest reconstruction. Simple Attention Non-AR provides a complete earlier baseline for latent-space analysis. Autoregressive MultiSlotting is selected as the main model for property-organization analysis because it gives the strongest raw and residual property-prediction results.

\begin{table}[t]
\centering
\caption{Reconstruction, generation, and interpolation benchmarks. Best available values are bolded.}
\label{tab:recon_gen_interp}
\resizebox{\textwidth}{!}{
\begin{tabular}{lccccccc}
\toprule
Model & Token acc. test/val & Seq. acc. test/val & Gen. validity & Uniqueness & Novelty & Interp. valid frac. & Family retention \\
\midrule
Linear Attention Non-AR 
& \textbf{0.999929} 
& \textbf{0.997671 }
& 0.9226 
& 0.8071 
& 0.9871 
& \textbf{1.0000} 
& 0.8225 \\
Simple Attention Non-AR 
& 0.998709 
& 0.961443 
& 0.8700 
& 0.7700 
& 0.9875 
& \textbf{1.0000} 
& 0.6926 \\
Autoregressive MultiSlotting  
& 0.9951 
& 0.9780 
& \textbf{1.0000} 
& \textbf{0.9964} 
& \textbf{0.9972} 
& \textbf{1.0000} 
& \textbf{0.8512} \\
\bottomrule
\end{tabular}
}
\end{table}

These results support a model-selection criterion based on latent organization rather than reconstruction alone. Linear Attention Non-AR shows that near-perfect reconstruction and valid decoding are achievable, but Autoregressive MultiSlotting provides the strongest evidence that chemically meaningful descriptors remain predictable after controlling for SELFIES sequence confounds. Traversal-based validation is analyzed in the following section.

\subsection{Confound-Aware Latent Property Prediction}

We evaluate whether latent property prediction reflects molecular structure rather than simple SELFIES statistics. The analysis asks: first, how well the latent space predicts RDKit descriptors; second, how strongly it encodes SELFIES confounds; and third, how much property signal remains after residualizing descriptors against these confounds.

Both non-autoregressive models strongly encode SELFIES statistics. This motivates residualized probing. HeavyAtomCount, MolWt, and BertzCT are strongly correlated with SELFIES length, with Spearman correlations of 0.9759, 0.9464, and 0.9042, respectively, as shown in Appendix \ref{app:correlations}. Raw probe performance on such descriptors is therefore not sufficient evidence of chemically meaningful latent organization.

\begin{table}[t]
\centering
\caption{Raw latent-to-property $R^2$ for key molecular descriptors. Best values are bolded.}
\label{tab:raw_r2}
\resizebox{\textwidth}{!}{
\begin{tabular}{lcccccccccccccccc}
\toprule
Model 
& HAC & MolWt & BertzCT & Ring & A.Ring & cLogP & TPSA & HBA & F.CSP3 & HBD & Spiro & Bridge & QED & NumRotBond & SA&Mean \\
\midrule
Linear Attention
& 0.9562 & 0.9136 & 0.9279 & 0.8549 & 0.8423 & 0.6677 & 0.6557 & 0.7071 & 0.7815 & 0.2171 & 0.0484 & 0.0737 & 0.3678 & 0.4734&0.4566&0.5963\\
Simple Attention
& 0.9622 & 0.9174 & 0.9305 & 0.8491 & 0.8498 & 0.6830 & 0.6994 & 0.7587 & 0.7842 & 0.2589 & 0.0509 & 0.0767 & 0.3465 &0.4844&0.4570&0.6072\\
AR MultiSlotting 
& \textbf{0.9750} & \textbf{0.9480} & \textbf{0.9512} & \textbf{0.9378} & \textbf{0.8937} & \textbf{0.7967} & \textbf{0.8059} & \textbf{0.8008} & \textbf{0.8951} & \textbf{0.4997} & \textbf{0.0878} & \textbf{0.1522} & \textbf{0.4624}& \textbf{0.6914}&\textbf{0.6412}&\textbf{0.7026} \\
\bottomrule
\end{tabular}
}
\end{table}

\begin{table}[t]
\centering
\caption{Residualized latent-to-property $R^2$ after removing SELFIES confounds. Best values are bolded.}
\label{tab:residual_r2}
\resizebox{\textwidth}{!}{
\begin{tabular}{lcccccccccccccccc}
\toprule
Model 
& HAC & MolWt & BertzCT & Ring & A.Ring & cLogP & TPSA & HBA & F.CSP3 & HBD & Spiro & Bridge & QED & NumRotBond &SA& Mean\\
\midrule
Linear Attention
& 0.3612 & 0.3799 & 0.5475 & 0.5143 & 0.6067 & 0.5406 & 0.4653 & 0.4735 & 0.6770 & 0.1782 & 0.0318 & 0.0472 & 0.3011 & 0.2065 &0.3887&0.3813\\
Simple Attention
& 0.3811 & 0.3109 & 0.5627 & 0.5151 & 0.6518 & 0.5770 & 0.5233 & 0.5404 & 0.7098 & 0.2202 & 0.0354 & 0.0443 & 0.2807 & 0.1992&0.3848&0.3958\\
AR MultiSlotting  
& \textbf{0.6554} & \textbf{0.5918} & \textbf{0.7009} & \textbf{0.7944} & \textbf{0.7559} & \textbf{0.6861} & \textbf{0.6863} & \textbf{0.6759} & \textbf{0.8313} & \textbf{0.4624} & \textbf{0.0728} & \textbf{0.1207} & \textbf{0.3907}& \textbf{0.4752} & \textbf{0.6015}&\textbf{0.5668}\\
\bottomrule
\end{tabular}
}
\end{table}

Autoregressive MultiSlotting gives the best raw and residual $R^2$ across the key descriptor panel (see Tables~\ref{tab:raw_r2} and \ref{tab:residual_r2}). For the main non-size descriptors, its residual performance remains high despite modest confound-to-property predictability: the confound-to-property $R^2$ values are 0.3612 for cLogP, 0.3786 for TPSA, 0.3805 for HBA, 0.3111 for FractionCSP3 and 0.0952 for SA-score. This supports the claim that these descriptors are not explained solely by SELFIES sequence statistics.

Table \ref{tab:descriptor_regimes} summarizes the descriptor regimes for the studied properties.

\begin{table}[t]
\centering
\caption{Descriptor regimes from the confound-aware probe analysis.}
\label{tab:descriptor_regimes}
\resizebox{\textwidth}{!}{
\begin{tabular}{lll}
\toprule
Regime & Descriptors & Evidence \\
\midrule
Confounded 
& HeavyAtomCount, MolWt, BertzCT, RingCount, AromaticRingCount 
& Strong correlation with SELFIES confounds; raw $R^2$ alone is not sufficient \\
Chemically meaningful residual signal 
& FractionCSP3, TPSA, cLogP, HBA, SA-score
& High AR raw $R^2$ and high AR residual $R^2$ after confound control \\
Weak or likely nonlinear 
& HBD, QED, NumRotatableBonds, NumSpiroAtoms, NumBridgeheadAtoms 
& Weaker or interaction-dependent linear signal; candidates for local nonlinear analysis \\
\bottomrule
\end{tabular}
}
\end{table}

\subsection{Linear Traversal Results}
\label{lineartraversal}
To test whether the learned linear probe directions correspond to controllable axes in latent space, we traversed along each regression direction and decoded the resulting latent points back to molecules. For each property, we repeated the experiment for \(N=50\) randomly sampled seed molecules. We summarize the response by the median trajectory together with the interquartile interval across seeds.

\begin{figure}[t]
    \centering

    \begin{subfigure}{0.40\textwidth}
        \centering
        \includegraphics[width=\linewidth]{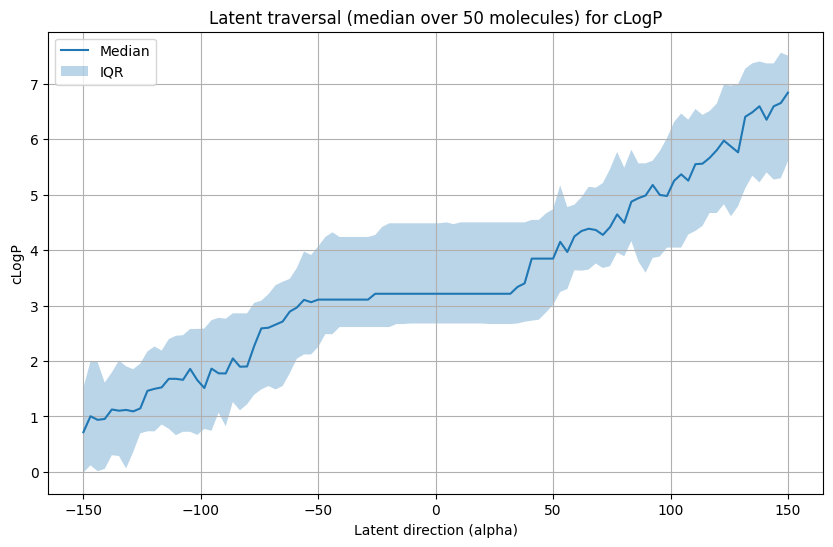}
    \end{subfigure}
    \hspace{0.04\textwidth}
    \begin{subfigure}{0.40\textwidth}
        \centering
        \includegraphics[width=\linewidth]{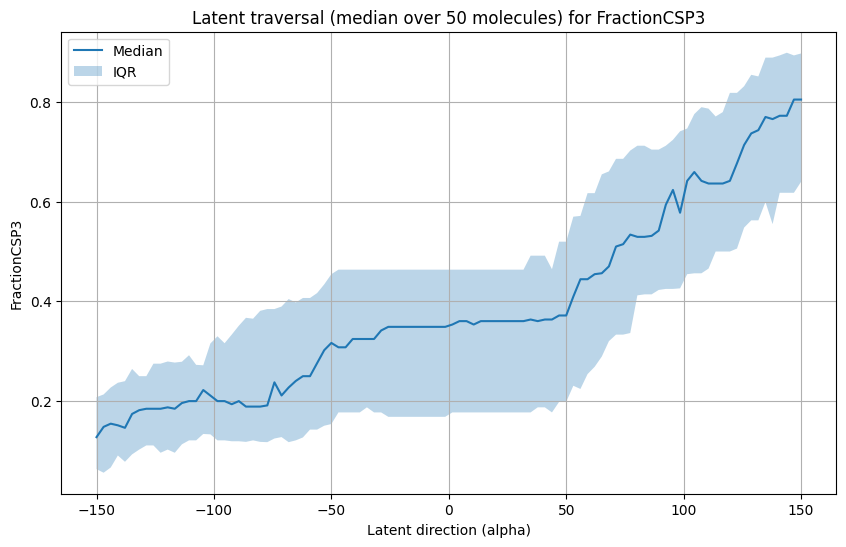}
    \end{subfigure}

    \vspace{0.4em}

    \begin{subfigure}{0.40\textwidth}
        \centering
        \includegraphics[width=\linewidth]{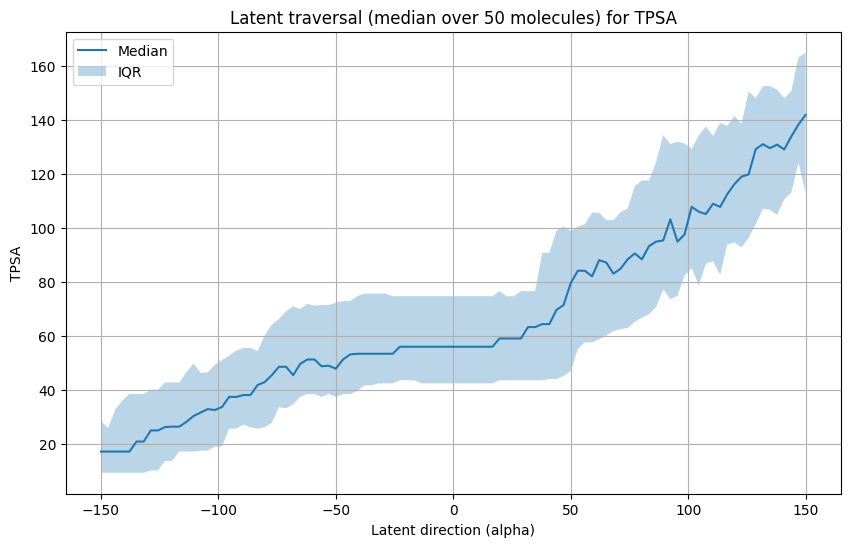}
    \end{subfigure}
    \hspace{0.04\textwidth}
    \begin{subfigure}{0.40\textwidth}
        \centering
        \includegraphics[width=\linewidth]{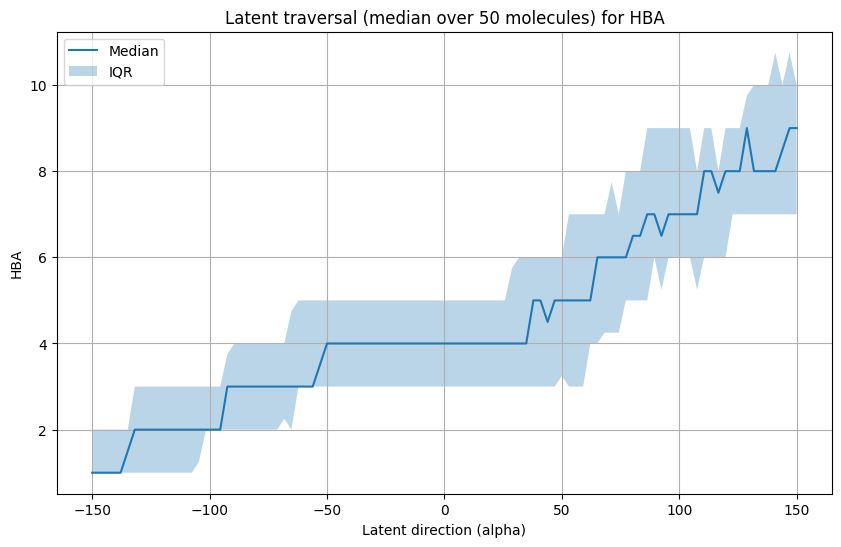}
    \end{subfigure}

    \caption{Latent traversals for cLogP, FractionCSP3, TPSA, and HBA reporting the median trajectory over $N=50$ randomly sampled seed molecules, with the interquartile interval across seeds.}
    \label{fig:linear-directions}
\end{figure}

Representative results are shown in Fig.~\ref{fig:linear-directions}. The four displayed properties —cLogP, FractionCSP3, TPSA, and HBA— all show clear monotonic increase along the positive traversal direction. FractionCSP3 spans nearly its full range, cLogP increases steadily, TPSA exhibits the strongest response, and HBA also follows a stable upward trend. These results show that the corresponding probe directions are not only predictive, but operational as steering directions after decoding.

Across the full traversal study, the clearest globally monotonic behavior was observed for HeavyAtomCount, cLogP, TPSA, HBA, FractionCSP3, and BertzCT. RingCount (including aromatic) was less regular, which is unsurprising given their discrete ring-based nature. SA-score was unstable under traversal: near-seed molecules remain easier to synthesize, while distant decoded molecules become harder. Importantly, some strongly confounded descriptors, notably HeavyAtomCount and BertzCT, still remained meaningfully traversable. This suggests that the confound structure captured by the model is not reducible to simple SELFIES length or token-count effects.

This point is important for interpreting the confound analysis. In earlier models, strongly confounded properties often lost most of their signal after residualization. In the present model, however, several confounded descriptors remain traversable, indicating that the latent organization is richer than a trivial encoding of sequence-level artifacts. To investigate this further, we report an inter-property correlation analysis in Appendix~\ref{app:interproperty-correlation}, examining alignment both between probe vectors and between traversal directions.

\subsection{Nonlinear Probe Results}
\subsubsection{Linear Probe versus MLP Performance}

We compare the autoregressive model with the simple-attention baseline on the expanded molecular-property set. The linear-attention baseline is excluded because preliminary traversal experiments showed decode collapse and repeated same-molecule outputs, making an MLP-versus-linear comparison uninformative for linear steerability; details are provided in Appendix~\ref{app:linearattention}.

The linear regression results have already shown that the autoregressive latent space is more predictive under linear readout. Here, we ask whether this advantage persists when the readout is allowed to be nonlinear. Using MLP probes, the autoregressive model remains better than the simple-attention baseline on every expanded property, both before and after confound control.

However, the gap between models decreases with the nonlinear readout (see the mean values in Tables \ref{tab:raw_r2}, \ref{tab:residual_r2} and \ref{tab:mlp_expanded_compare}). On raw targets, the mean $R^2$ gap decreases from $0.096$ with linear probes ($0.703$ vs.\ $0.607$) to $0.064$ with MLP probes ($0.895$ vs.\ $0.831$). After residualization, the gap decreases from $0.171$ with linear probes ($0.567$ vs.\ $0.396$) to $0.100$ with MLP probes ($0.859$ vs.\ $0.759$).

This pattern suggests that both latent spaces contain nonlinear local information about the descriptors, but that the autoregressive model organizes this information in a more globally linear form. This distinction is important because property-directed traversal depends on the existence of stable linear directions, not merely on nonlinear decodability.

\begin{table}[t]
\centering
\caption{MLP probe performance on the expanded molecular-property set. The autoregressive model remains more predictive than the simple-attention baseline under nonlinear readout across both raw and residualized evaluations.}
\label{tab:mlp_expanded_compare}
\resizebox{\textwidth}{!}{
\begin{tabular}{l*{16}{c}}
\toprule
Model / target & HAC & MolWt & BertzCT & Ring & A.Ring & cLogP & TPSA & HBA & F.CSP3 & HBD & Spiro & Bridge & QED & NumRotBond &SA& Mean\\
\midrule
AR MLP raw
& \textbf{0.996} & \textbf{0.990} & \textbf{0.986} & \textbf{0.985} & \textbf{0.955} & \textbf{0.934} & \textbf{0.946} & \textbf{0.926} & \textbf{0.982} & \textbf{0.792} & \textbf{0.628} & \textbf{0.785} & \textbf{0.803} & \textbf{0.909} & \textbf{0.814} & \textbf{0.895} \\

Simple-attention MLP raw
& 0.990 & 0.979 & 0.975 & 0.953 & 0.933 & 0.890 & 0.905 & 0.905 & 0.959 & 0.644 & 0.436 & 0.680 & 0.685 & 0.821 & 0.711 & 0.831 \\
\midrule
AR MLP residual
& \textbf{0.933} & \textbf{0.938} & \textbf{0.917} & \textbf{0.958} & \textbf{0.902} & \textbf{0.901} & \textbf{0.916} & \textbf{0.884} & \textbf{0.966} & \textbf{0.778} & \textbf{0.608} & \textbf{0.771} & \textbf{0.771} & \textbf{0.849} & \textbf{0.795} & \textbf{0.859} \\

Simple-attention MLP residual
& 0.830 & 0.834 & 0.851 & 0.855 & 0.857 & 0.837 & 0.842 & 0.834 & 0.930 & 0.626 & 0.423 & 0.656 & 0.640 & 0.701 & 0.671 & 0.759 \\
\bottomrule
\end{tabular}
}
\end{table}

\subsubsection{Properties with Strong Nonlinear Structure}

The MLP comparison shows that nonlinear readouts recover additional descriptor information from the autoregressive latent space. However, this does not imply that the corresponding properties are suitable for linear traversal. A property may be decodable from \(z\) while still lacking a single global direction along which it changes monotonically.

We therefore focus on descriptors for which the autoregressive model shows a large gap between linear and MLP probe performance. These cases indicate latent signal that is present, but organized in a nonlinear or interaction-dependent manner. Table~\ref{tab:ar_nonlinear_properties} reports the clearest examples.

\begin{table}[H]
\centering
\caption{Autoregressive descriptors with strong nonlinear structure. Large differences between linear and MLP probes indicate that the property is encoded in the latent space but not primarily through a single global linear direction.}
\label{tab:ar_nonlinear_properties}
\resizebox{\textwidth}{!}{
\begin{tabular}{lccccc}
\toprule
Probe / property & HBD & NumRotatableBonds & NumSpiroAtoms & NumBridgeheadAtoms & QED \\
\midrule
AR Linear Regression raw   & 0.500 & 0.691 & 0.088 & 0.152 & 0.462 \\
AR MLP raw      & 0.792 & 0.909 & 0.628 & 0.785 & 0.803 \\
AR Linear Regression resid & 0.462 & 0.475 & 0.073 & 0.121 & 0.391 \\
AR MLP resid    & 0.778 & 0.849 & 0.608 & 0.771 & 0.771 \\
\bottomrule
\end{tabular}
}
\end{table}

NumSpiroAtoms and NumBridgeheadAtoms are the strongest examples: linear probes are nearly ineffective, whereas MLP probes recover substantial predictive power in both the raw and residual settings. HBD, QED and NumRotatableBonds show the same pattern more moderately. These results suggest that such descriptors are represented in the latent space, but through nonlinear combinations of latent features rather than through a stable global axis.
Additional insight about $\Delta R^2$ and full per-property results are reported in Appendix~\ref{app:delta_r2}.

\section{Conclusion, limitations and future directions}

We presented a confound-aware framework for evaluating whether an unsupervised SELFIES-based Transformer-VAE learns chemically meaningful and steerable latent structure. The main result is that the autoregressive MultiSlotting Transformer-VAE organizes several RDKit descriptors in directions that are not only predictable, but also operationally steerable. The nonlinear-probe analysis further shows that some descriptors remain latently encoded even when they are not well captured by a single global linear direction.

The study is limited to computed RDKit descriptors and therefore does not establish performance on experimental biochemical, pharmacokinetic, toxicity, docking, or synthesis outcomes. In addition, the confound panel captures only a minimal set of SELFIES-level shortcuts, and the steering results are evaluated for one model family and one large SELFIES dataset. Broader validation across datasets, architectures, molecular domains, and decoding strategies remains necessary.

Future work should extend confound control to graph-size, scaffold, functional-group, and dataset-frequency effects; combine linear steering with chemical trust-region constraints such as fingerprint similarity, scaffold retention, synthetic accessibility, and multi-objective trade-offs; and explore local nonlinear steering methods based on MLP gradients or learned latent vector fields.



\bibliographystyle{unsrtnat}
\bibliography{molecula_state_of_art_refs}

@article{weininger1988smiles,
  title = {{SMILES}, a chemical language and information system. 1. Introduction to methodology and encoding rules},
  author = {Weininger, David},
  journal = {Journal of Chemical Information and Computer Sciences},
  volume = {28},
  number = {1},
  pages = {31--36},
  year = {1988},
  doi = {10.1021/ci00057a005}
}

@article{bjerrum2017smiles,
  title = {{SMILES} enumeration as data augmentation for neural network modeling of molecules},
  author = {Bjerrum, Esben Jannik},
  journal = {arXiv preprint arXiv:1703.07076},
  year = {2017},
  eprint = {1703.07076},
  archivePrefix = {arXiv}
}

@article{krenn2020selfies,
  title = {Self-referencing embedded strings ({SELFIES}): A 100\% robust molecular string representation},
  author = {Krenn, Mario and Hase, Florian and Nigam, AkshatKumar and Friederich, Pascal and Aspuru-Guzik, Alan},
  journal = {Machine Learning: Science and Technology},
  volume = {1},
  number = {4},
  pages = {045024},
  year = {2020},
  doi = {10.1088/2632-2153/aba947}
}

@inproceedings{vaswani2017attention,
  title = {Attention is All You Need},
  author = {Vaswani, Ashish and Shazeer, Noam and Parmar, Niki and Uszkoreit, Jakob and Jones, Llion and Gomez, Aidan N. and Kaiser, Lukasz and Polosukhin, Illia},
  booktitle = {Advances in Neural Information Processing Systems},
  volume = {30},
  year = {2017}
}

@inproceedings{devlin2019bert,
  title = {{BERT}: Pre-training of Deep Bidirectional Transformers for Language Understanding},
  author = {Devlin, Jacob and Chang, Ming-Wei and Lee, Kenton and Toutanova, Kristina},
  booktitle = {Proceedings of the 2019 Conference of the North American Chapter of the Association for Computational Linguistics: Human Language Technologies},
  pages = {4171--4186},
  year = {2019},
  publisher = {Association for Computational Linguistics},
  doi = {10.18653/v1/N19-1423}
}

@article{rives2021biological,
  title = {Biological structure and function emerge from scaling unsupervised learning to 250 million protein sequences},
  author = {Rives, Alexander and Meier, Joshua and Sercu, Tom and Goyal, Siddharth and Lin, Zeming and Liu, Jason and Guo, Demi and Ott, Myle and Zitnick, C. Lawrence and Ma, Jerry and Fergus, Rob},
  journal = {Proceedings of the National Academy of Sciences},
  volume = {118},
  number = {15},
  pages = {e2016239118},
  year = {2021},
  doi = {10.1073/pnas.2016239118}
}

@article{honda2019smiles,
  title = {{SMILES} Transformer: Pre-trained Molecular Fingerprint for Low Data Drug Discovery},
  author = {Honda, Shion and Shi, Shoi and Ueda, Hiroki R.},
  journal = {arXiv preprint arXiv:1911.04738},
  year = {2019},
  eprint = {1911.04738},
  archivePrefix = {arXiv},
  doi = {10.48550/arXiv.1911.04738}
}

@article{chithrananda2020chemberta,
  title = {{ChemBERTa}: Large-Scale Self-Supervised Pretraining for Molecular Property Prediction},
  author = {Chithrananda, Seyone and Grand, Gabriel and Ramsundar, Bharath},
  journal = {arXiv preprint arXiv:2010.09885},
  year = {2020},
  eprint = {2010.09885},
  archivePrefix = {arXiv},
  doi = {10.48550/arXiv.2010.09885}
}

@article{yuksel2023selformer,
  title = {{SELFormer}: Molecular Representation Learning via {SELFIES} Language Models},
  author = {Yuksel, Atakan and Ulusoy, Erva and Unlu, Atabey and Dogan, Tunca},
  journal = {arXiv preprint arXiv:2304.04662},
  year = {2023},
  eprint = {2304.04662},
  archivePrefix = {arXiv},
  doi = {10.48550/arXiv.2304.04662}
}

@article{bagal2021molgpt,
  title = {{MolGPT}: Molecular Generation Using a Transformer-Decoder Model},
  author = {Bagal, Viraj and Aggarwal, Rishal and Vinod, P. K. and Priyakumar, U. Deva},
  journal = {Journal of Chemical Information and Modeling},
  volume = {62},
  number = {9},
  pages = {2064--2076},
  year = {2021},
  doi = {10.1021/acs.jcim.1c00600}
}

@article{sanchezlengeling2018inverse,
  title = {Inverse molecular design using machine learning: Generative models for matter engineering},
  author = {Sanchez-Lengeling, Benjamin and Aspuru-Guzik, Alan},
  journal = {Science},
  volume = {361},
  number = {6400},
  pages = {360--365},
  year = {2018},
  doi = {10.1126/science.aat2663}
}

@article{vamathevan2019applications,
  title = {Applications of machine learning in drug discovery and development},
  author = {Vamathevan, Jessica and Clark, Dominic and Czodrowski, Paul and Dunham, Ian and Ferran, Edgardo and Lee, George and Li, Bin and Madabhushi, Anant and Shah, Pankaj and Spitzer, Michaela and Zhao, Shanrong},
  journal = {Nature Reviews Drug Discovery},
  volume = {18},
  number = {6},
  pages = {463--477},
  year = {2019},
  doi = {10.1038/s41573-019-0024-5}
}

@article{du2022molgensurvey,
  title = {{MolGenSurvey}: A systematic survey in machine learning models for molecule design},
  author = {Du, Yuanqi and Fu, Tianfan and Sun, Jimeng and Liu, Shengchao},
  journal = {arXiv preprint arXiv:2203.14500},
  year = {2022},
  eprint = {2203.14500},
  archivePrefix = {arXiv}
}

@inproceedings{you2018gcpn,
  title = {Graph Convolutional Policy Network for Goal-Directed Molecular Graph Generation},
  author = {You, Jiaxuan and Liu, Bowen and Ying, Rex and Pande, Vijay and Leskovec, Jure},
  booktitle = {Advances in Neural Information Processing Systems},
  volume = {31},
  year = {2018}
}

@article{zhou2019optimization,
  title = {Optimization of molecules via deep reinforcement learning},
  author = {Zhou, Zhou and Kearnes, Steven and Li, Li and Zare, Richard N. and Riley, Patrick},
  journal = {Scientific Reports},
  volume = {9},
  pages = {10752},
  year = {2019},
  doi = {10.1038/s41598-019-47148-x}
}

@article{jensen2019graph,
  title = {A graph-based genetic algorithm and generative model/{Monte} {Carlo} tree search for the exploration of chemical space},
  author = {Jensen, Jan H.},
  journal = {Chemical Science},
  volume = {10},
  number = {12},
  pages = {3567--3572},
  year = {2019},
  doi = {10.1039/C8SC05372C}
}

@inproceedings{fu2022reinforced,
  title = {Reinforced Genetic Algorithm for Structure-Based Drug Design},
  author = {Fu, Tianfan and Gao, Wenhao and Coley, Connor W. and Sun, Jimeng},
  booktitle = {Advances in Neural Information Processing Systems},
  volume = {35},
  pages = {12325--12338},
  year = {2022}
}

@article{gomezbombarelli2018automatic,
  title = {Automatic chemical design using a data-driven continuous representation of molecules},
  author = {Gomez-Bombarelli, Rafael and Wei, Jennifer N. and Duvenaud, David and Hernandez-Lobato, Jose Miguel and Sanchez-Lengeling, Benjamin and Sheberla, Dennis and Aguilera-Iparraguirre, Jorge and Hirzel, Timothy D. and Adams, Ryan P. and Aspuru-Guzik, Alan},
  journal = {ACS Central Science},
  volume = {4},
  number = {2},
  pages = {268--276},
  year = {2018},
  doi = {10.1021/acscentsci.7b00572}
}

@inproceedings{kingma2013autoencoding,
  title = {Auto-Encoding Variational Bayes},
  author = {Kingma, Diederik P. and Welling, Max},
  booktitle = {International Conference on Learning Representations},
  year = {2014},
  eprint = {1312.6114},
  archivePrefix = {arXiv}
}

@inproceedings{jin2018junction,
  title = {Junction Tree Variational Autoencoder for Molecular Graph Generation},
  author = {Jin, Wengong and Barzilay, Regina and Jaakkola, Tommi},
  booktitle = {Proceedings of the 35th International Conference on Machine Learning},
  series = {Proceedings of Machine Learning Research},
  volume = {80},
  pages = {2323--2332},
  year = {2018},
  publisher = {PMLR}
}

@article{cao2018molgan,
  title = {{MolGAN}: An implicit generative model for small molecular graphs},
  author = {De Cao, Nicola and Kipf, Thomas},
  journal = {arXiv preprint arXiv:1805.11973},
  year = {2018},
  eprint = {1805.11973},
  archivePrefix = {arXiv}
}

@inproceedings{zang2020moflow,
  title = {{MoFlow}: An invertible flow model for generating molecular graphs},
  author = {Zang, Chengxi and Wang, Fei},
  booktitle = {Proceedings of the 26th ACM SIGKDD International Conference on Knowledge Discovery and Data Mining},
  pages = {617--626},
  year = {2020},
  doi = {10.1145/3394486.3403104}
}

@inproceedings{vignac2023digress,
  title = {{DiGress}: Discrete Denoising Diffusion for Graph Generation},
  author = {Vignac, Clement and Krawczuk, Igor and Siraudin, Antoine and Wang, Bohan and Cevher, Volkan and Frossard, Pascal},
  booktitle = {International Conference on Learning Representations},
  year = {2023}
}

@article{griffiths2020constrained,
  title = {Constrained Bayesian optimization for automatic chemical design using variational autoencoders},
  author = {Griffiths, Ryan-Rhys and Hernandez-Lobato, Jose Miguel},
  journal = {Chemical Science},
  volume = {11},
  number = {2},
  pages = {577--586},
  year = {2020},
  doi = {10.1039/C9SC04026A}
}

@inproceedings{eckmann2022limo,
  title = {{LIMO}: Latent Inceptionism for Targeted Molecule Generation},
  author = {Eckmann, Peter and Sun, Kunyang and Zhao, Bo and Feng, Mudong and Gilson, Michael K. and Yu, Rose},
  booktitle = {Proceedings of the 39th International Conference on Machine Learning},
  series = {Proceedings of Machine Learning Research},
  volume = {162},
  pages = {5777--5792},
  year = {2022},
  publisher = {PMLR}
}

@article{du2023chemspace,
  title = {{ChemSpace}: Interpretable and Interactive Chemical Space Exploration},
  author = {Du, Yuanqi and Liu, Xinhao and Shah, Neil M. and Liu, Shengchao and Zhang, Jie and Zhou, Bolei},
  journal = {Transactions on Machine Learning Research},
  year = {2023}
}

@inproceedings{wei2024chemflow,
  title = {Navigating Chemical Space with Latent Flows},
  author = {Wei, Guanghao and Huang, Yining and Duan, Chenru and Song, Yue and Du, Yuanqi},
  booktitle = {Advances in Neural Information Processing Systems},
  volume = {37},
  year = {2024}
}

@article{alain2016understanding,
  title = {Understanding intermediate layers using linear classifier probes},
  author = {Alain, Guillaume and Bengio, Yoshua},
  journal = {arXiv preprint arXiv:1610.01644},
  year = {2016},
  eprint = {1610.01644},
  archivePrefix = {arXiv}
}

@inproceedings{hewitt2019designing,
  title = {Designing and Interpreting Probes with Control Tasks},
  author = {Hewitt, John and Liang, Percy},
  booktitle = {Proceedings of the 2019 Conference on Empirical Methods in Natural Language Processing and the 9th International Joint Conference on Natural Language Processing},
  pages = {2733--2743},
  year = {2019},
  publisher = {Association for Computational Linguistics},
  doi = {10.18653/v1/D19-1275}
}

@article{belinkov2022probing,
  title = {Probing Classifiers: Promises, Shortcomings, and Advances},
  author = {Belinkov, Yonatan},
  journal = {Computational Linguistics},
  volume = {48},
  number = {1},
  pages = {207--219},
  year = {2022},
  doi = {10.1162/coli_a_00422}
}

@article{alberts2024unraveling,
  title={Unraveling molecular structure: A multimodal spectroscopic dataset for chemistry},
  author={Alberts, Marvin and Schilter, Oliver and Zipoli, Federico and Hartrampf, Nina and Laino, Teodoro},
  journal={Advances in Neural Information Processing Systems},
  volume={37},
  pages={125780--125808},
  year={2024}
}

\newpage

\appendix
\section*{Appendix}
\addcontentsline{toc}{section}{Appendix}

\section{Code, dataset, and additional data info}

\subsection{Experimental Setup and Reproducibility}
\label{app:experimental_setup}

All code, notebooks, model definitions, checkpoints, and curated artifacts are available in the anonymized repository:
\url{https://anonymous.4open.science/r/MolecuLA-7A68/}.
All paths are repository-relative. The released checkpoints are treated as fixed pretrained inputs, while downstream artifacts can be regenerated from \texttt{data/}, \texttt{checkpoints/}, \texttt{models/}, and \texttt{study/notebooks/}. The large dataset CSV and checkpoints are stored as Git LFS assets; after cloning, \texttt{git lfs pull} should be run before executing notebooks.

\paragraph{Randomness and determinism.}
Unless otherwise specified, all experiments use \texttt{SEED = 42}. Dataset splits use \texttt{train\_test\_split(..., random\_state=42, shuffle=True)} for both split stages. The notebooks seed Python \texttt{random}, NumPy, \texttt{torch.manual\_seed}, and \texttt{torch.cuda.manual\_seed\_all} when CUDA is available; explicit generators use \texttt{np.random.default\_rng(42)} or documented offsets from this seed.

Exact floating-point values may vary slightly across PyTorch, CUDA, cuDNN, BLAS, GPU model, and RDKit versions. The intended reproducibility contract is therefore fixed dataset rows, tokenizer, splits, checkpoint weights, model settings, probe grids, random seeds, and output schemas, with small numerical differences expected across machines.

\paragraph{Dataset, tokenization, and splits.}
The molecular dataset is derived from the dataset introduced by \citet{alberts2024unraveling}. The original dataset contains molecular structures together with multiple experimental and predicted molecular-property modalities, including NMR- and mass-spectrometry-related annotations. In this study, we use only the molecular structure information: the SMILES column is extracted and converted to SELFIES for model training and evaluation. The processed SMILES/SELFIES dataset used in this work is included in the anonymized repository and is sufficient to reproduce all experiments reported here.

The processed dataset is loaded from \texttt{data/smiles\_selfies\_full.csv} and must contain at least \texttt{smiles} and \texttt{selfies} columns. The released tokenizer is \texttt{data/selfies\_tokenizer.json}, with \texttt{max\_len = 77} and \texttt{vocab\_size = 110} for the non-autoregressive checkpoints. SELFIES strings are encoded by prepending \texttt{<SOS>}, appending \texttt{<EOS>}, and padding with \texttt{<PAD>}.

The autoregressive notebook constructs its own vocabulary with \texttt{<PAD>}, \texttt{<SOS>}, \texttt{<EOS>}, and \texttt{MASK}; the corresponding checkpoint records \texttt{vocab\_size = 111}. This difference is intentional and is handled by the autoregressive model path.

The standard split is 80\% train, 10\% validation, and 10\% test. For the released dataset this gives 635,522 training rows, 79,440 validation rows, and 79,441 test rows. RDKit-derived panels retain invalid molecules in the table but mark them with \texttt{is\_rdkit\_valid}; probe fitting and property summaries use valid molecules with finite values for the relevant target.

\paragraph{Model settings.}
The model registry in \texttt{models/registry.py} is the source of truth for the public model configurations.

\begin{table}[H]
\centering
\caption{Transformer-VAE model settings used in the released checkpoints.}
\label{tab:app_model_settings}
\resizebox{\textwidth}{!}{
\begin{tabular}{lcccccccc}
\toprule
Model & Hidden size & Latent size & Max length & Heads & Slots & Encoder/decoder depth & Checkpoint vocab & Checkpoint epoch \\
\midrule
\texttt{linear\_attention} & 256 & 512 & 77 & 8 & 8 & 1 shared non-AR layer & 110 & 50 \\
\texttt{simple\_attention} & 256 & 256 & 77 & 8 & 8 & 1 shared non-AR layer & 110 & 50 \\
\texttt{autoregressive} & 256 & 256 & 77 & 8 & 8 & 3 encoder layers, 2 decoder layers & 111 & 50 \\
\bottomrule
\end{tabular}
}
\end{table}

The checkpoints contain model weights, optimizer state, training history, checkpoint epoch, and vocabulary size. The optimizer is AdamW with learning rate \texttt{3e-4}, betas \texttt{(0.9, 0.999)}, epsilon \texttt{1e-8}, and weight decay \texttt{1e-2}. All three architectures use Transformer-style attention blocks with dropout $p=0.1$ inside the feed-forward blocks. The non-autoregressive blocks use feed-forward width $4 \times$ hidden size, while the autoregressive blocks use $2 \times$ hidden size.

\begin{table}[H]
\centering
\caption{Loss terms used by the three Transformer-VAE variants.}
\label{tab:app_model_losses}
\resizebox{\textwidth}{!}{
\begin{tabular}{llll}
\toprule
Model & Reconstruction term & KL term & Length term \\
\midrule
\texttt{linear\_attention} & token cross entropy over non-pad targets & default $\beta = 0.01$ & MSE on predicted length, $\alpha = 0.1$ \\
\texttt{simple\_attention} & token cross entropy over non-pad targets & default $\beta = 0.01$ & MSE on predicted length, $\alpha = 1.0$ \\
\texttt{autoregressive} & token cross entropy with \texttt{ignore\_index=<PAD>} & default $\beta = 0.01$, overridden by schedule & none \\
\bottomrule
\end{tabular}
}
\end{table}

\paragraph{Training.}
The notebook \texttt{study/notebooks/00\_data\_tokenization\_and\_training.ipynb} contains the rerunnable training workflow. Training was performed on a single NVIDIA A100 40GB GPU and took approximately 2 hours 30 minutes for 50 epochs. The released checkpoints can be used directly without retraining, but rerunning the notebook with the same settings regenerates the models from the same tokenization and split logic.

\begin{table}[H]
\centering
\caption{Autoregressive training settings.}
\label{tab:app_ar_training_settings}
\resizebox{0.72\textwidth}{!}{
\begin{tabular}{lc}
\toprule
Setting & Value \\
\midrule
epochs & 50 \\
batch size & 512 \\
hidden size & 256 \\
latent size & 256 \\
attention heads & 8 \\
number of slots & 8 \\
encoder layers & 3 \\
decoder layers & 2 \\
optimizer & AdamW \\
learning rate & \texttt{3e-4} \\
betas & \texttt{(0.9, 0.999)} \\
epsilon & \texttt{1e-8} \\
weight decay & \texttt{1e-2} \\
initial \texttt{beta} variable & \texttt{0.01} \\
effective beta schedule & \texttt{(epoch \% cycle\_length / cycle\_length) * max\_beta} \\
\texttt{max\_beta} & \texttt{0.03} \\
\texttt{cycle\_length} & 15 \\
corruption & \texttt{False} \\
training loader shuffle & \texttt{True} \\
validation loader shuffle & \texttt{False} \\
notebook \texttt{num\_workers} & 1 \\
checkpoint selection & best validation sequence accuracy \\
\bottomrule
\end{tabular}
}
\end{table}

\paragraph{Latent encoding and model-quality benchmarks.}
Latents are encoded with the model in evaluation mode and without gradient tracking. The helper in \texttt{study/common/latents.py} returns the encoder mean $\mu$ as \texttt{float32}; its default batch size is 256. The main benchmark notebooks use \texttt{SEED = 42}, batch size 1024 for reconstruction, prior-sampling, and interpolation evaluations, and \texttt{N\_STEPS = 11} for interpolation tests. Family-retention tests use seed 42 and a minimum family group size of 20.

Quality outputs include token accuracy, sequence accuracy, reconstruction validity, prior validity, interpolation validity, novelty and uniqueness summaries where available, and family-retention summaries. The curated result folders contain compact CSV/JSON summaries and figures only; regenerated latent arrays and full decode caches are excluded.

\paragraph{Linear probes, residualization, and Ridge directions.}
The notebook \texttt{study/notebooks/02\_linear\_probes\_panels\_and\_residuals.ipynb} is the canonical workflow for the linear probe analysis. The main reported latent-to-property results, especially for the two ablation transformers, use ordinary linear regression to measure how much property signal is available through a single global linear readout. For each target, finite rows are selected, latent coordinates are standardized using train rows, target values are standardized using train rows, and train/validation/test $R^2$ scores are reported.

Ridge models are used in auxiliary places where regularization is useful, especially for residualization, confound prediction, and some direction-fitting utilities. Residualization fits a multivariate $C \rightarrow Y$ Ridge model on train rows, predicts all valid rows, and stores \texttt{resid\_\{property\}} columns. The residual probe then repeats the same linear-probe procedure on these residual targets.

\begin{table}[H]
\centering
\caption{Linear-probe and residualization settings.}
\label{tab:app_linear_probe_settings}
\resizebox{0.78\textwidth}{!}{
\begin{tabular}{lc}
\toprule
Setting & Value \\
\midrule
\texttt{SEED} & 42 \\
latent encoding \texttt{BATCH\_SIZE} & 1024 \\
split random state & 42 \\
input scaling & \texttt{StandardScaler} fit on train rows \\
target scaling & \texttt{StandardScaler} fit on train rows \\
main reported probe & ordinary linear regression \\
reported metrics & train/validation/test $R^2$ \\
Ridge alpha grid, where used & \texttt{np.logspace(-3, 3, 13)} \\
residualization model & multivariate Ridge, $C \rightarrow Y$ \\
\bottomrule
\end{tabular}
}
\end{table}

Autoregressive Step 3 source notebooks used fixed Ridge settings for some compact recovered tables and direction utilities: \texttt{alpha=1e-2} for $Z \rightarrow Y$ and $Z \rightarrow Y_{\mathrm{residual}}$, \texttt{alpha=1e1} for $C \rightarrow Y$ residualization, and \texttt{alpha=1.0} for $Z \rightarrow C$ confound directions.

\paragraph{MLP probe settings.}
The notebook \texttt{study/notebooks/04\_mlp\_probes\_all\_properties.ipynb} is the canonical MLP-probe workflow. It compares nonlinear MLP probes against the linear baselines on raw and residualized targets.

\begin{table}[H]
\centering
\caption{Canonical MLP probe settings.}
\label{tab:app_mlp_probe_settings}
\resizebox{0.75\textwidth}{!}{
\begin{tabular}{lc}
\toprule
Setting & Value \\
\midrule
\texttt{SEED} & 42 \\
\texttt{BATCH\_SIZE} & 8192 \\
\texttt{MAX\_EPOCHS} & 15 \\
\texttt{MIN\_EPOCHS} & 6 \\
\texttt{PATIENCE} & 4 \\
minimum validation improvement & $10^{-3}$ in validation $R^2$ \\
learning rate & $10^{-3}$ \\
weight decay & $10^{-4}$ \\
hidden width & 256 \\
architecture & Linear, ReLU, Linear, ReLU, Linear \\
optimizer & AdamW \\
loss & MSE on standardized target \\
input scaling & latent train mean/std; std below $10^{-8}$ replaced by 1.0 \\
target scaling & train target mean/std; std below $10^{-8}$ replaced by 1.0 \\
AMP & enabled when CUDA is available \\
\bottomrule
\end{tabular}
}
\end{table}

The MLP notebook stores the best in-memory state by validation $R^2$. Training stops after at least \texttt{MIN\_EPOCHS} once \texttt{PATIENCE} consecutive epochs fail to improve validation $R^2$ by more than $10^{-3}$. Reported scores are computed after inverse-transforming predictions back to original target units.

The autoregressive source MLP notebooks used the same batch size, maximum epochs, patience, learning rate, weight decay, and hidden width, but the compact recovered MLP tables come from an architecture \texttt{Linear -> GELU -> LayerNorm -> Linear -> GELU -> Linear}. These source notebooks early-stopped on validation-loss improvement greater than $10^{-5}$, with patience 4.

\paragraph{Latent traversal and direction settings.}
The traversal analysis uses raw-property directions fitted with ordinary linear regression. For each target property, latent means $z$ and property values $y$ are split with \texttt{test\_size=0.2} and \texttt{random\_state=42}. A \texttt{StandardScaler}+\texttt{LinearRegression} pipeline is fitted on the training split. The learned coefficient vector is converted back to the original latent coordinate scale by dividing by the scaler standard deviation, then normalized to unit norm before traversal.

Two traversal protocols are used. The dense single-origin traversal is used for cLogP and starts from the latent origin $z_0=0$, traversing
\[
z(\alpha) = z_0 + \alpha \hat{w},
\qquad
\alpha \in [-200, 200],
\]
with 5000 evenly spaced steps. The multi-seed traversal is used for the main property panels. It samples 50 dataset latents as seeds and traverses each seed along the fitted raw-property direction with 100 evenly spaced steps over
\[
\alpha \in [-150, 150].
\]
Decoded molecules are scored with RDKit, and the reported traversal curves show the median property value across the 50 seeds, with the interquartile range shown as a band. This protocol is applied to F.CSP3, TPSA, cLogP, HBA, MolWt, BertzCT, HAC, Ring, and A.Ring.

\begin{table}[H]
\centering
\caption{Latent traversal settings used in the source traversal notebook.}
\label{tab:app_traversal_settings}
\resizebox{0.82\textwidth}{!}{
\begin{tabular}{lc}
\toprule
Setting & Value \\
\midrule
direction model & \texttt{StandardScaler} + ordinary \texttt{LinearRegression} \\
direction target & raw property value \\
train/test split & \texttt{test\_size=0.2}, \texttt{random\_state=42} \\
direction rescaling & divide coefficient vector by latent scaler std \\
direction normalization & unit $\ell_2$ norm \\
dense traversal seed & latent origin, $z_0=0$ \\
dense traversal property & cLogP \\
dense traversal alphas & \texttt{np.linspace(-200.0, 200.0, 5000)} \\
multi-seed traversal seeds & 50 dataset latents \\
multi-seed traversal alphas & \texttt{np.linspace(-150.0, 150.0, 100)} \\
multi-seed traversal properties & F.CSP3, TPSA, cLogP, HBA, MolWt, BertzCT, HAC, Ring, A.Ring \\
decode batch size & 4096 for multi-seed traversal decoding \\
reported traversal summary & median with interquartile range across seeds \\
\bottomrule
\end{tabular}
}
\end{table}

\paragraph{Output regeneration policy.}
The notebooks regenerate large panels, latent matrices, direction arrays, MLP weights, and decode caches locally. The repository commits only compact artifacts needed for inspection: summary CSV files, JSON manifests, selected figures, checkpoints, and data assets tracked by Git LFS.

Curated evidence is stored under \texttt{study/results/}. This includes benchmark and family-retention metrics, confound summaries, compact $R^2$/correlation tables, direction-quality tables, single-property traversal summaries, representative figures, MLP-vs-linear probe summaries, and ablation comparison summaries. Large regenerated assets such as full panels, latent arrays, \texttt{.npz} direction caches, duplicate large CSV files, and MLP weight files are excluded.

\subsection{Property Panel and Descriptor Tiers}
\label{app:property_panel}

This appendix details the molecular property panel used for latent-space probing listed in Table~\ref{tab:property_panel_tiers}. All properties are computed from RDKit molecular graphs after decoding or standardizing the corresponding molecules; they are not used as supervision during VAE training. We organize the descriptors into three tiers according to interpretability and expected sensitivity to representation-level confounds.

\paragraph{Tier 1: simple size and count descriptors.}
Tier 1 contains descriptors that are chemically interpretable but often strongly correlated with coarse SELFIES statistics such as sequence length, token count, or molecular size. These properties are useful sanity checks for whether the latent space encodes basic molecular structure, but high predictability may partly reflect representation-level scale effects.

\paragraph{Tier 2: physicochemical and structural descriptors.}
Tier 2 contains descriptors that remain graph-computable but encode more specific chemical information. These properties probe polarity, lipophilicity, saturation, flexibility, rigidity, and local three-dimensional structure. They are more informative for latent direction discovery because they are less reducible to a single notion of molecular size.

\paragraph{Tier 3: composite and graph-complexity descriptors.}
Tier 3 contains higher-level descriptors that combine multiple structural or physicochemical factors. These targets are more demanding probes of the latent space because they summarize non-local graph organization or aggregate several simpler molecular properties into a single scalar score.

\begin{table}[H]
\centering
\caption{
Molecular descriptors used in the property panel, grouped according to the tier system used in the analysis.
}
\label{tab:property_panel_tiers}
\small
\begin{tabular}{lll}
\toprule
Tier & Descriptor & Chemical interpretation \\
\midrule
1 & MolWt 
& Molecular size and mass \\

1 & HeavyAtomCount 
& Number of non-hydrogen atoms \\

1 & HBD 
& Hydrogen-bond donor count \\

1 & HBA 
& Hydrogen-bond acceptor count \\

1 & RingCount 
& Number of rings \\

1 & AromaticRingCount 
& Number of aromatic rings \\

1 & NumRotatableBonds 
& Molecular flexibility \\

\midrule
2 & cLogP 
& Crippen lipophilicity estimate \\

2 & TPSA 
& Topological polar surface area \\

2 & FractionCSP3 
& Carbon saturation and three-dimensionality \\

2 & NumSpiroAtoms 
& Spiro-center count and rigid 3D ring junctions \\

2 & NumBridgeheadAtoms 
& Ring fusion complexity and rigidity \\

\midrule
3 & BertzCT 
& Graph-topological complexity \\

3 & SA score 
& Estimated synthetic accessibility \\

3 & QED 
& Quantitative estimate of drug-likeness \\
\bottomrule
\end{tabular}
\end{table}

Tier 1 descriptors primarily test whether the latent representation captures coarse molecular scale and simple count information. Tier 2 descriptors probe more chemically specific axes, such as polarity, lipophilicity, saturation, and ring-system rigidity. Tier 3 descriptors are the most composite: BertzCT summarizes graph complexity, SA score combines fragment rarity with structural penalties, and QED aggregates several physicochemical desirability terms. This tiering is used only for interpretation; all descriptors are treated uniformly as scalar targets in the probing experiments.

\section{VAE architecture and benchmarking}

\subsection{Transformer-VAE Architecture Details}
\label{app:architecture}

The Transformer-VAE operates on SELFIES-tokenized molecules \(x=(x_1,\ldots,x_T)\), where \(T\) denotes the sequence length and \(x_t\) is the token at position \(t\). Token embeddings are combined with positional encodings and processed by Transformer encoder blocks, producing contextual token representations
\[
H = \mathrm{Enc}_{\theta}(x) \in \mathbb{R}^{T \times h},
\]
where \(\mathrm{Enc}_{\theta}\) is the Transformer encoder with parameters \(\theta\), \(h\) is the hidden dimension, and \(H=(h_1,\ldots,h_T)\) contains the encoded token states. Multi-head self-attention enables each non-padding token to attend to all other tokens, allowing the encoder to capture non-local molecular dependencies such as ring closures, branches, and functional groups that may be separated in the SELFIES string.

Rather than compressing the sequence through a single pooled vector, the encoder uses learned multi-slot pooling. A set of learned slot queries \(\{q_k\}_{k=1}^{K_s}\), where \(K_s\) is the number of slots, attends over the encoded token states:
\[
s_k
=
\sum_{t=1}^{T}
\alpha_{kt} V h_t,
\qquad
\alpha_{kt}
=
\operatorname{softmax}_{t}
\left(
q_k^\top K h_t
\right),
\]
where \(s_k\) denotes the \(k\)-th latent slot, \(q_k\) is its learned query vector, \(K\) and \(V\) are learned key and value projection matrices, and \(\alpha_{kt}\) is the attention weight assigned by slot \(k\) to token position \(t\). The softmax is taken over the token index \(t\). This allows different slots to specialize to different aspects of the molecular representation, which is useful when molecular properties are distributed across multiple structural motifs.

Each slot is mapped to a Gaussian posterior component:
\[
\mu_k = W_{\mu}s_k,
\qquad
\log \sigma_k^2 = W_{\sigma}s_k,
\]
where \(\mu_k\) and \(\sigma_k^2\) are the mean and diagonal variance of the \(k\)-th slot-level Gaussian component, and \(W_{\mu}\) and \(W_{\sigma}\) are learned linear projections.

The slot-level Gaussians are combined into a global posterior using confidence-weighted latent pooling. Slots with lower predicted uncertainty receive larger weights:
\[
\alpha_k
=
\operatorname{softmax}_{k}
\left(
\frac{q^\top K\mu_k}{\tau}
+
\lambda c_k
\right),
\qquad
c_k
=
-\log
\sum_j
\exp
\left(
\log \sigma_{k,j}^2
\right).
\]
Here, \(\alpha_k\) is the normalized weight assigned to slot \(k\), \(q\) is a learned global pooling query, \(K\) is a learned projection matrix applied to the slot mean, \(\tau>0\) is a temperature parameter, \(\lambda\) controls the influence of the confidence term, and \(c_k\) is a scalar confidence score for slot \(k\). The index \(j\) runs over latent dimensions, and \(\sigma_{k,j}^2\) is the variance of the \(j\)-th latent dimension for slot \(k\). Since \(c_k\) decreases with the total predicted variance, slots with lower uncertainty receive larger confidence scores.

The final posterior parameters are
\[
\mu
=
\sum_k
\alpha_k \mu_k,
\qquad
\sigma^2
=
\sum_k
\alpha_k \sigma_k^2,
\]
where \(\mu\) and \(\sigma^2\) denote the mean and diagonal variance of the global approximate posterior. This yields
\[
q_{\theta}(z \mid x)
=
\mathcal{N}
\left(
\mu,
\operatorname{diag}(\sigma^2)
\right),
\]
where \(z\) is the latent molecular representation and \(q_{\theta}(z \mid x)\) is the approximate posterior distribution parameterized by the encoder.

The latent vector is sampled using the reparameterization trick:
\[
z
=
\mu
+
\sigma \odot \epsilon,
\qquad
\epsilon \sim \mathcal{N}(0,I),
\]
where \(\epsilon\) is standard Gaussian noise, \(I\) is the identity matrix, \(\sigma\) is the elementwise standard deviation, and \(\odot\) denotes elementwise multiplication.

The decoder is an autoregressive Transformer decoder. The latent vector is projected into a memory representation,
\[
m = W_m z,
\]
where \(W_m\) is a learned projection matrix and \(m\) is the latent memory used by the decoder. The decoder reconstructs the SELFIES sequence as
\[
p_{\theta}(x \mid z)
=
\prod_{t=1}^{T}
p_{\theta}
\left(
x_t
\mid
x_{<t}, z
\right),
\]
where \(x_{<t}\) denotes the previously generated tokens before position \(t\), and \(p_{\theta}(x \mid z)\) is the decoder likelihood parameterized by \(\theta\). Training uses teacher forcing with shifted ground-truth inputs. Causal self-attention prevents access to future tokens, while cross-attention allows each decoding position to condition on the latent memory \(m\). This prevents direct encoder--decoder token copying and ensures that reconstruction depends on the global latent representation.

\subsection{Interpolation Smoothness and Family-Conditioned Retention}
\label{app:interpolation_smoothness}

We evaluate latent-space interpolation as an additional diagnostic of the learned molecular geometry. 
Given two validation molecules $x_1$ and $x_2$ with latent codes
\[
z_1 = E(x_1), 
\qquad 
z_2 = E(x_2),
\]
we construct the linear latent path
\[
z(t) = (1-t)z_1 + tz_2, 
\qquad t \in [0,1].
\]
Each intermediate point is decoded and mapped back to a molecular structure. Unless otherwise stated, interpolation paths use $K=11$ decoded steps and statistics are averaged over $N=200$ sampled endpoint pairs.

\begin{figure}[t]
    \centering
    \includegraphics[width=0.82\textwidth]{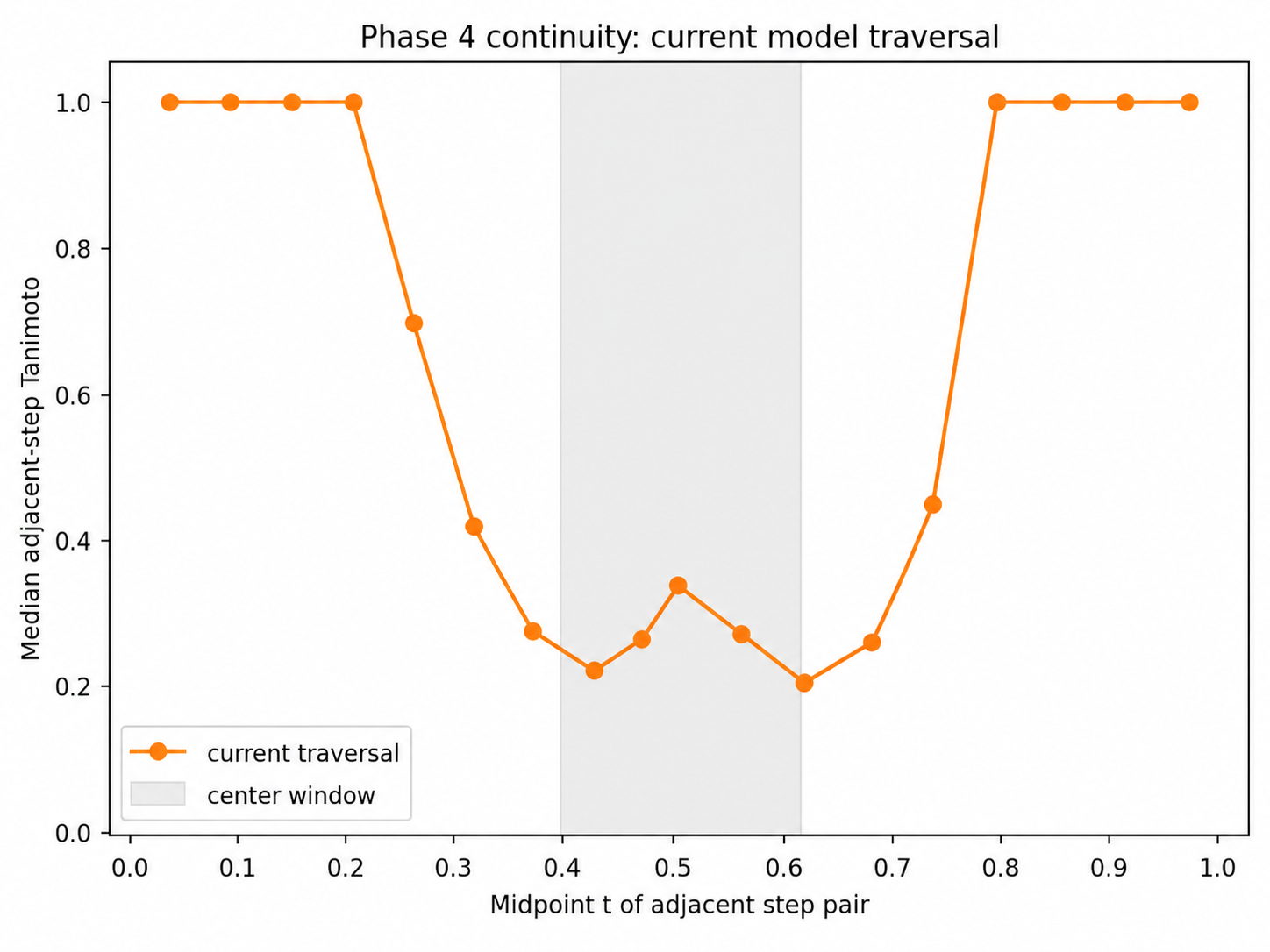}
    \caption{
    Interpolation continuity for the current model. The curve reports the median adjacent-step Tanimoto similarity over $N=200$ interpolation paths. The shaded region marks the central interpolation window, where structural transitions are largest.
    }
    \label{fig:app_phase4_continuity}
\end{figure}

We report two interpolation diagnostics. First, we measure decoded validity at each interpolation step. The model obtains a valid fraction of $1.0$ across all interpolation positions, indicating that the decoder remains chemically valid along linear paths between encoded validation molecules. Second, we measure local structural continuity using adjacent-step Tanimoto similarity:
\[
s_i =
\operatorname{Tanimoto}
\left(
\operatorname{FP}(\hat{x}(t_i)),
\operatorname{FP}(\hat{x}(t_{i+1}))
\right),
\]
where $\hat{x}(t_i)$ is the decoded molecule at interpolation step $t_i$ and $\operatorname{FP}(\cdot)$ denotes the molecular fingerprint. Figure~\ref{fig:app_phase4_continuity} reports the median adjacent-step Tanimoto similarity as a function of the midpoint between adjacent interpolation steps. The model is most continuous near the endpoints, while the central region exhibits lower similarity, indicating larger structural transitions near the midpoint despite preserved validity.

We further evaluate whether interpolation preserves local semantic structure by performing \emph{family-conditioned interpolation}. In this setting, both endpoints are sampled from the same functional family $f$, defined by deterministic SMARTS matching on the validation set. For each decoded interpolation point, we compute the retention indicator
\[
r_f(t_i)
=
\mathbf{1}
\left[
\hat{x}(t_i) \in f
\right],
\]
and summarize each family by
\[
R_f
=
\frac{1}{K}
\sum_{i=1}^{K}
r_f(t_i).
\]
Thus, $R_f$ measures the fraction of decoded interpolation steps that remain within the target functional family. This diagnostic separates chemical validity from motif-level semantic continuity: a path may remain valid while losing the functional group shared by its endpoints.

The functional groups used in this analysis are listed in Table~\ref{tab:app_functional_family_smarts}. They cover common local motifs including alcohols, phenols, ethers, amines, amides, carboxylic acids, esters, aldehydes, ketones, nitriles, and sulfonamides.

\begin{table}[t]
\centering
\caption{
Functional-family definitions used for family-conditioned interpolation. Families are assigned by SMARTS matching on decoded molecules.
}
\label{tab:app_functional_family_smarts}
\begin{tabular}{lll}
\toprule
Family & Chemical motif & SMARTS pattern \\
\midrule
Alcohol & $R{-}OH$ & \texttt{[OX2H][CX4]} \\
Phenol & $Ar{-}OH$ & \texttt{[OX2H][c]} \\
Ether & $R{-}O{-}R'$ & \texttt{[OD2]([{\#}6;!\$(C=O)])[{\#}6;!\$(C=O)]} \\
Amine & primary/secondary/tertiary amine & \texttt{[NX3;H2,H1,H0;!\$(N[C,S,P]=O)]} \\
Amide & $R{-}C(=O){-}NR_2$ & \texttt{[NX3][CX3](=[OX1])[{\#}6,{\#}7,{\#}8]} \\
Carboxylic acid & $R{-}C(=O)OH$ & \texttt{[CX3](=O)[OX2H1]} \\
Ester & $R{-}C(=O)OR'$ & \texttt{[CX3](=O)[OX2H0][{\#}6]} \\
Aldehyde & $R{-}CHO$ & \texttt{[CX3H1](=O)[{\#}6]} \\
Ketone & $R{-}C(=O){-}R'$ & \texttt{[{\#}6][CX3](=O)[{\#}6]} \\
Nitrile & $R{-}C{\equiv}N$ & \texttt{[CX2]{\#}N} \\
Sulfonamide & $R{-}SO_2{-}NR_2$ & \texttt{[SX4](=[OX1])(=[OX1])([{\#}6])[NX3]} \\
\bottomrule
\end{tabular}
\end{table}

\begin{figure}[t]
    \centering
    \includegraphics[width=0.82\textwidth]{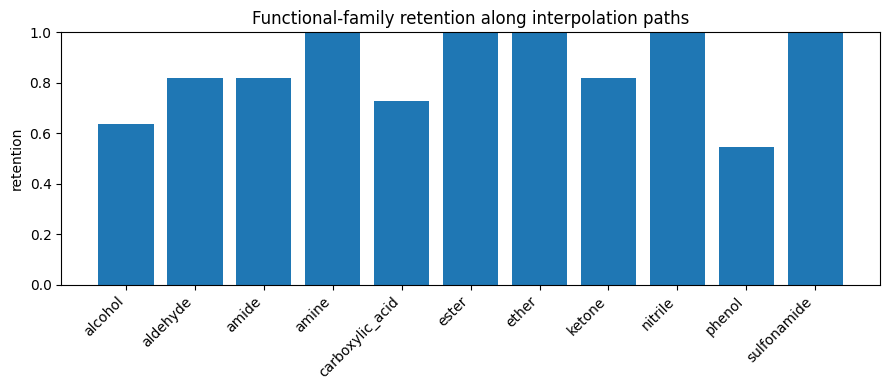}
    \caption{
    Functional-family retention along family-conditioned interpolation paths. Bars report the average fraction of decoded interpolation steps that remain within the endpoint functional family. All paths have valid fraction $1.0$; the variation shown here therefore reflects semantic motif retention rather than decoding validity.
    }
    \label{fig:app_family_retention}
\end{figure}

Figure~\ref{fig:app_family_retention} reports the resulting family-retention scores. All family-conditioned paths retain perfect decoded validity. Motif retention, however, varies by functional group. Amines, esters, ethers, nitriles, and sulfonamides are retained across all decoded interpolation steps. Aldehydes, amides, and ketones also show high retention. Alcohols, carboxylic acids, and phenols are less stable, with phenols showing the lowest retention. This suggests that some functional motifs correspond to more locally connected regions in latent space, whereas others are more easily disrupted by interpolation.

\section{Robustness analysis of the latent space}
Because held-out $R^2$ alone does not distinguish genuine latent signal from statistical artifact, we complement probe performance with a set of robustness controls. The goal is to test whether predictive performance is reproducible, whether it collapses under label destruction, and whether the inferred directions depend only on the latent information itself rather than on an arbitrary coordinate system.
\subsection{Robustness Tests conducted }

First, we assess \emph{bootstrap stability}. For each property, let $w$ denote the probe direction fit on the full training set, and let $w^{(b)}$ denote the direction obtained from bootstrap resample $b \in \{1,\dots,B\}$. Stability is summarized by the sign-aligned cosine similarity
$
s^{(b)} = \frac{ \langle w^{(b)}, w \rangle }{\|w^{(b)}\|_2 \, \|w\|_2}.
$, \(\langle \cdot,\cdot\rangle\) denotes the Euclidean inner product.

High agreement indicates that the recovered direction is not a fragile consequence of a particular sample realization, but a reproducible feature of the latent representation.

Second, we apply a \emph{permutation control}. The latent codes are kept fixed, but the property values are randomly permuted within the training split,
$
y_i^{\pi} = y_{\pi(i)},
$
before refitting the probe. If the original $R^2$ reflects genuine latent--property association, predictive performance should collapse under permutation on validation and test data. This control directly checks that the measured signal is not produced by chance fitting capacity or by regularization artifacts.

Finally, we test \emph{rotation and basis invariance}. Linear probe directions should represent geometric structure in latent space rather than dependence on a particular coordinate basis. Concretely, if the latent codes are transformed by an orthogonal map $Q$, so that
$
z_i' = Q z_i, Q^\top Q = I,
$
then an equivalent predictor exists in the rotated basis:
\[
\hat y_i = w^\top z_i + b = (Qw)^\top z_i' + b,
\]
and held-out predictions should remain unchanged. This verifies that the interpretation is attached to the latent subspace geometry, not to individual coordinates.

A property is therefore treated as robust only when its predictive signal is reproducible across resamples, disappears under permutation, and remains consistent under admissible changes of latent parameterization.

\subsection{Bootstrap stability of learned directions}
\begin{figure}[t]
    \centering
    \includegraphics[width=\textwidth]{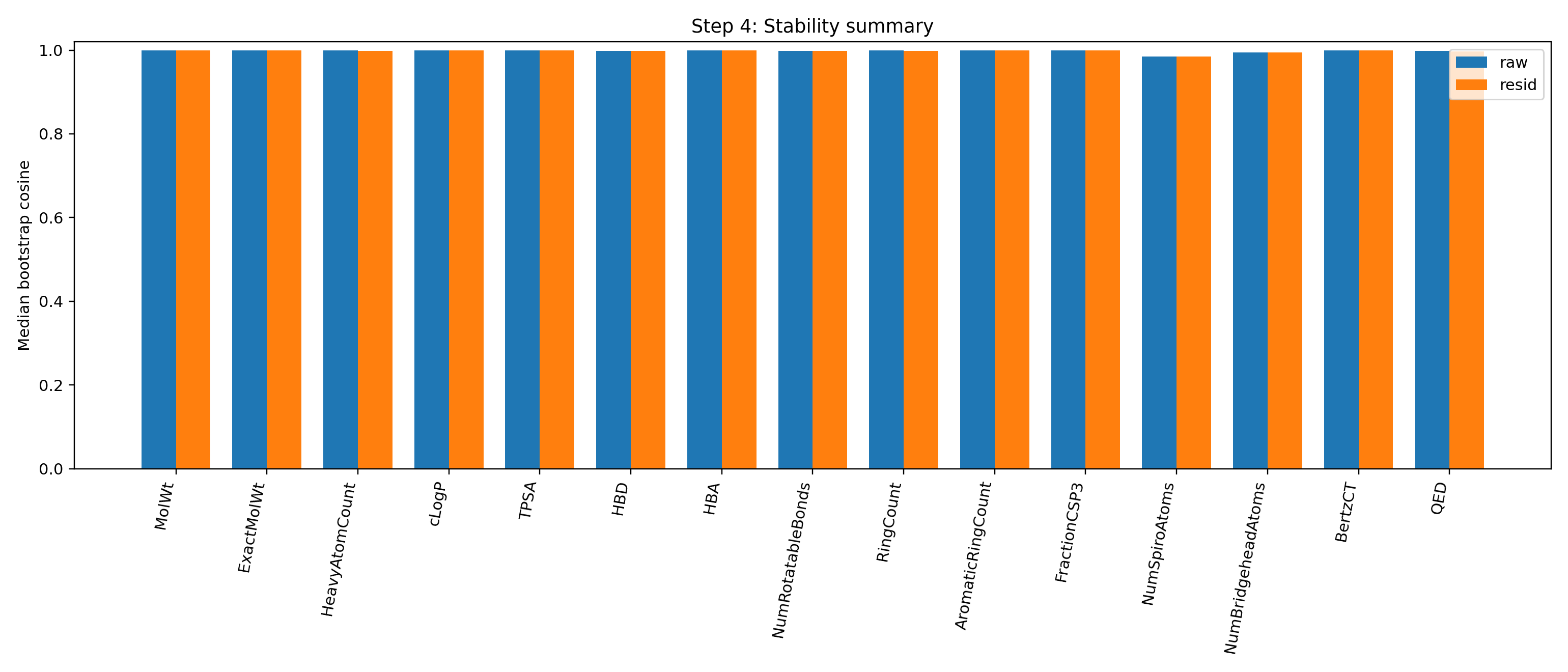}
    \caption{Bootstrap stability of probe directions. Bars report the median cosine similarity between the reference direction and the directions obtained by refitting on bootstrap resamples of the training data, for raw and residual targets.}
    \label{fig:bootstrap_stability}
\end{figure}

Figure~\ref{fig:bootstrap_stability} summarizes the bootstrap stability analysis.
For each property, the probe was refit across multiple bootstrap resamples of the training set (B = 100) , and the learned directions were compared to the reference direction using cosine similarity.
The plotted quantity is therefore a directional stability measure rather than a direct performance metric.

For both raw and residual targets, the median bootstrap cosine is close to $1$ for nearly all properties.
This indicates that the learned directions are highly reproducible under resampling and are not driven by a small subset of training examples or by instability in the fitting procedure.
Importantly, this stability remains high after residualization, showing that the residual-property directions are still well defined in latent space.
However, bootstrap stability alone should not be interpreted as evidence that a direction is free of confound influence; rather, it shows that the recovered direction is robust.

\subsection{Permutation and random-direction controls}
\begin{figure}[ht]
    \centering
    \includegraphics[width=\textwidth]{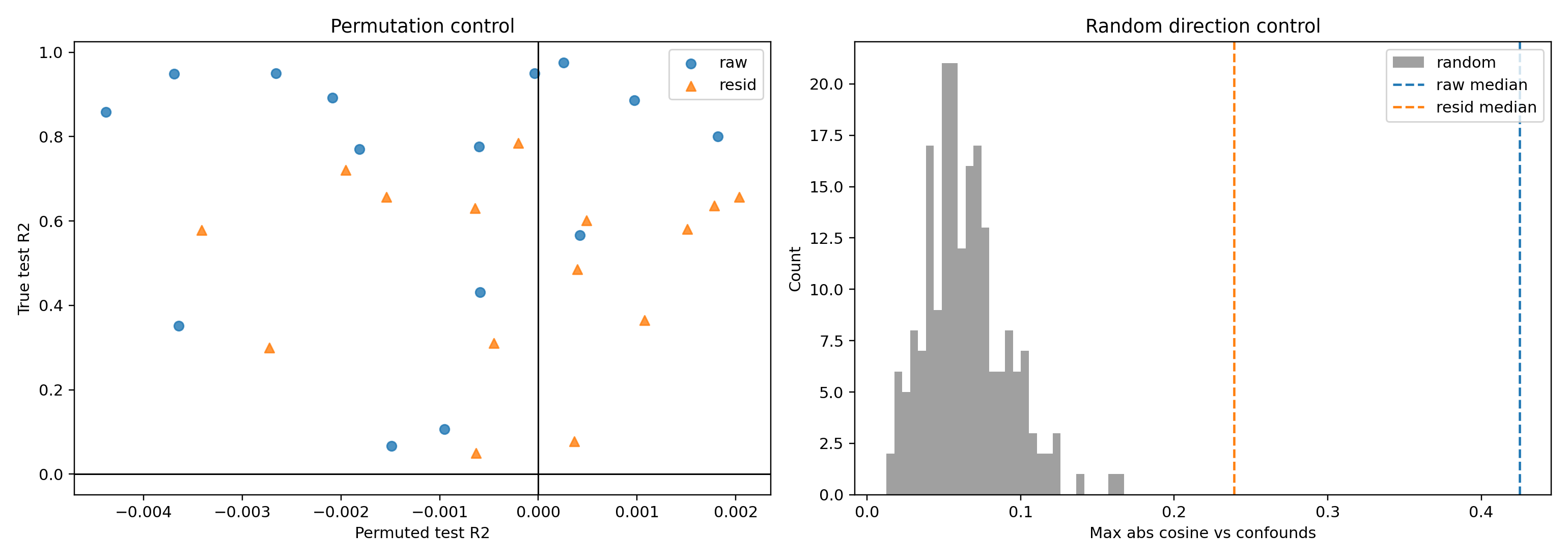}
    \caption{Control analyses. \textbf{Left:} permutation control, showing that test-set $R^2$ becomes approximately null when the training labels are permuted. \textbf{Right:} random-direction control, comparing the observed maximum absolute cosine similarity to confound directions against a null distribution from random latent directions.}
    \label{fig:controls}
\end{figure}

Figure~\ref{fig:controls} reports two control analyses.
The left panel shows the permutation control, in which training labels are randomly permuted before fitting the probe.
Under permutation, test-set $R^2$ values collapse to approximately zero, with some slightly negative values, as expected when the true association between latent representations and labels has been destroyed.
This indicates that the non-zero $R^2$ values observed with the true labels do not arise from trivial artifacts of the training procedure or data split.

The right panel compares the observed alignment between property directions and confound directions to a null distribution obtained from random latent directions.
For each random direction, the maximum absolute cosine similarity to the confound directions is computed.
The random baseline remains concentrated at relatively low values.
The median raw-property alignment lies far above this null distribution, showing that raw property directions are substantially more aligned with confound directions than would be expected by chance.
Residualization reduces this alignment, but the median residual-property alignment remains above the random baseline.
Thus, residualization suppresses confound-driven signal, but does not render the learned directions fully orthogonal to the confound subspace.

\section{Additional confound analysis }
\label{app:confounds_robustness}

To complement the main results, we report additional analyses assessing
(i) the degree to which chemical properties are associated with simple SELFIES-derived confounds and 
(ii) the geometric alignment between property directions and confound directions in latent space.

\subsection{Correlations between properties and confounds}
\label{app:correlations}

\begin{figure}[ht]

\centering
\begin{subfigure}[t]{0.48\textwidth}
    \centering
    \includegraphics[width=0.78\textwidth]{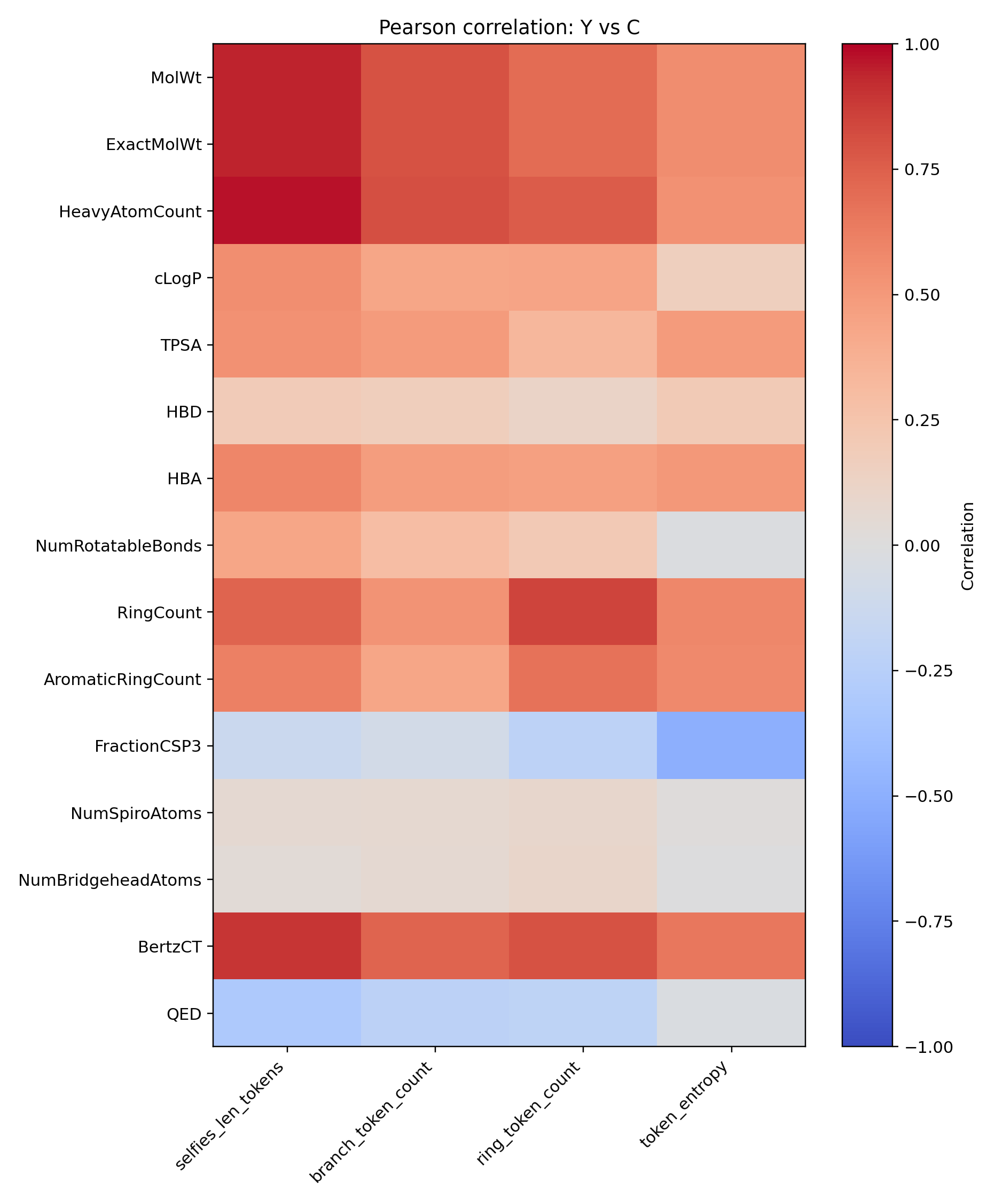}
\end{subfigure}
\hfill
\begin{subfigure}[t]{0.48\textwidth}
    \centering
    \includegraphics[width=0.78\textwidth]{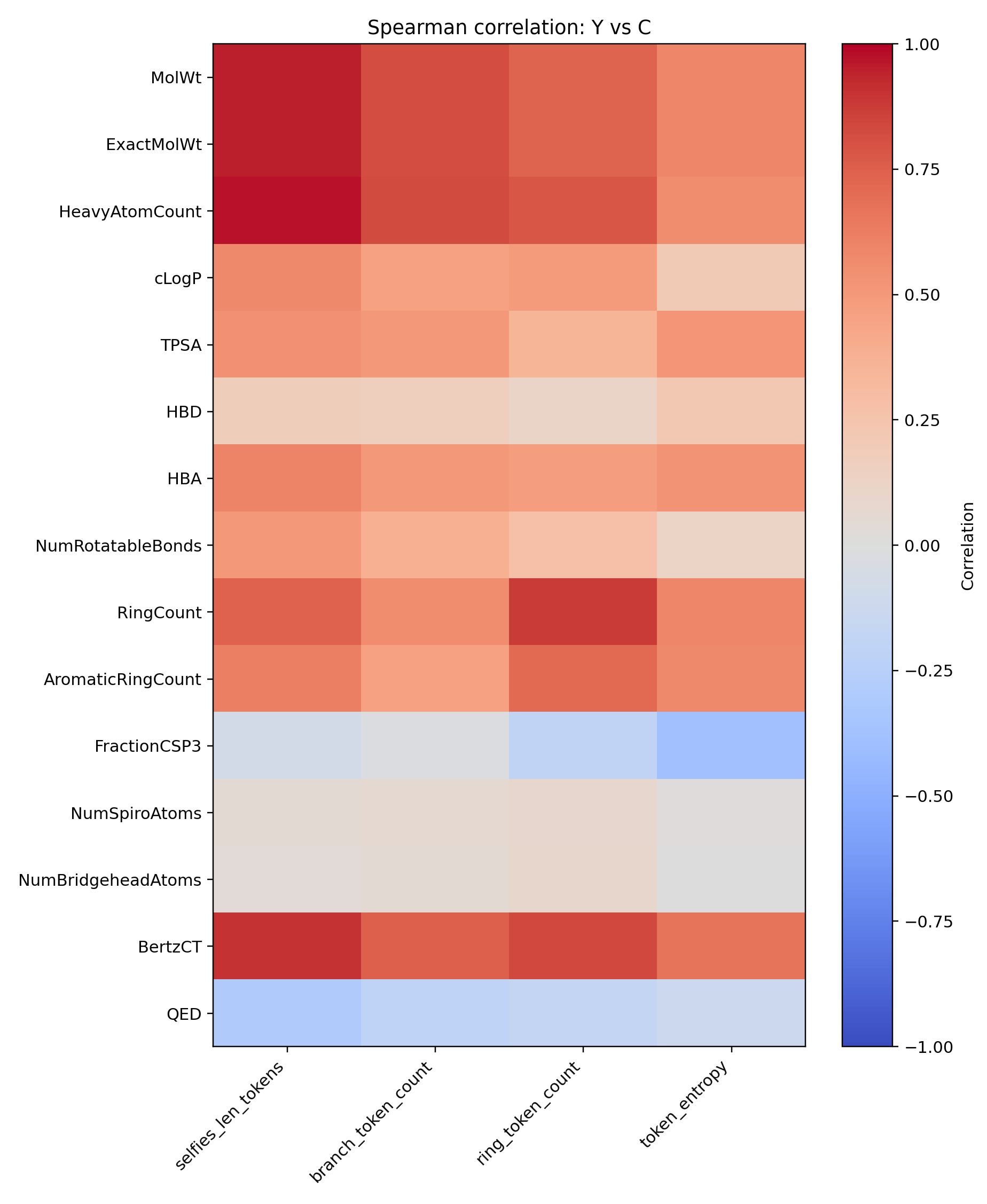}
\end{subfigure}
\caption{Correlation between molecular properties and SELFIES-derived confounds. Pearson and Spearman correlations show highly consistent structure, indicating robust monotonic associations between several properties and simple sequence-level statistics.}
    \label{fig:correlation_confounds}
\end{figure}

Figure~\ref{fig:correlation_confounds} reports Pearson and Spearman correlations between the chemical properties and the confounds. The two correlation matrices are highly consistent, indicating that the observed associations are not driven solely by linear outliers, but reflect robust monotonic structure.

The strongest correlations are observed for size- and complexity-related properties.
In particular, \texttt{MolWt}, \texttt{HeavyAtomCount}, and \texttt{BertzCT} are strongly positively correlated with SELFIES length and, to a lesser extent, with branch- and ring-token counts.
Similarly, \texttt{RingCount} and \texttt{AromaticRingCount} show clear positive association with the ring-token count, as expected.
These patterns indicate that part of the predictive signal for such properties can be recovered from relatively shallow sequence-level statistics alone.

Several additional properties, including \texttt{cLogP}, \texttt{TPSA}, \texttt{HBA}, \texttt{HBD}, and \texttt{NumRotatableBonds}, also exhibit moderate correlations with one or more confounds, suggesting partial confound mediation.
By contrast, \texttt{NumSpiroAtoms} and \texttt{NumBridgeheadAtoms} remain weakly correlated overall, while \texttt{FractionCSP3} and \texttt{QED} show weaker or negative associations, especially with entropy- and length-related variables.
Overall, these results support the use of residualized targets as a stricter test of whether the latent representation encodes property information beyond simple SELFIES-derived statistics.

\subsection{Cosine similarity between property and confound directions}
To assess the geometry of the learned directions, we compute the cosine similarity between each property direction $w_j$ and each confound direction $v_k$,
\begin{equation}
\cos(w_j, v_k) = \frac{w_j^\top v_k}{\lVert w_j \rVert \, \lVert v_k \rVert}.
\end{equation}

\begin{figure}[t]
    \centering
    \includegraphics[width=\textwidth]{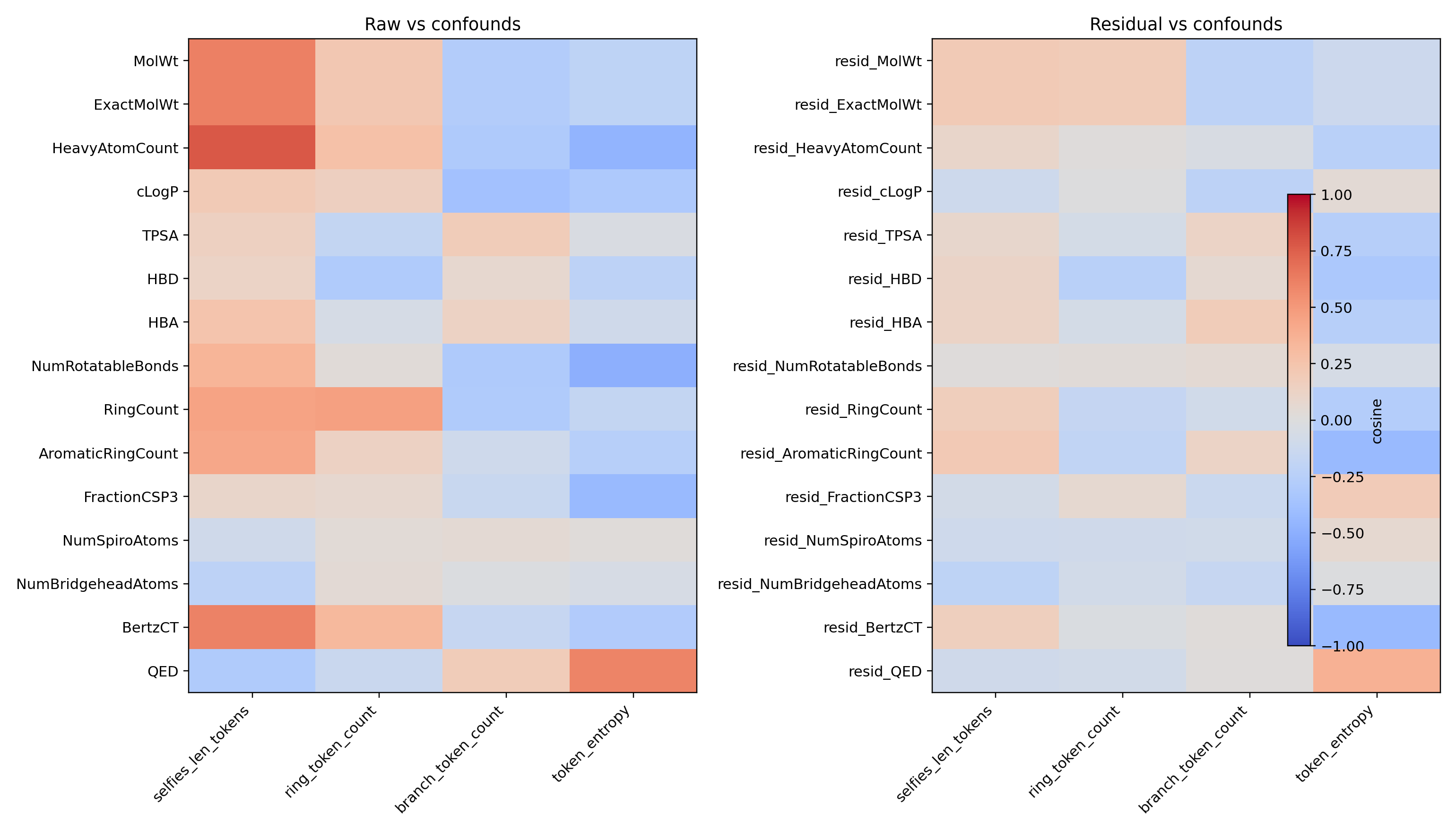}
    \caption{Cosine similarity between property directions and confound directions in latent space. \textbf{Left:} raw-property directions. \textbf{Right:} residual-property directions. Residualization reduces, but does not entirely eliminate, directional overlap with the confound subspace.}
    \label{fig:cosine_heatmaps}
\end{figure}

Figure~\ref{fig:cosine_heatmaps} shows these cosine similarities for both raw-property directions and residual-property directions.
The raw heatmap reveals broad alignment between many property directions and confound directions.
This is particularly pronounced for \texttt{MolWt},  \texttt{HeavyAtomCount}, and \texttt{BertzCT}, which are strongly aligned with the SELFIES length direction, and for ring-related descriptors, which align with ring-token count.
These results are consistent with the correlation analyses and indicate that part of the raw latent predictability is mediated by latent axes encoding simple sequence-level statistics.

After residualization, the cosine structure is markedly attenuated and closer to zero overall.
Nevertheless, several residual-property directions still retain non-negligible alignment with one or more confound directions.
This suggests that residualization removes the component of the target that is statistically explained by the measured confounds, but does not enforce complete geometric disentanglement in latent space.
In other words, residual probes can retain genuine property-related information beyond the explicit confounds, while still occupying partially overlapping latent subspaces.

\section{Additional Insights for Linear Probing and Monotonic Traversal}
\label{app:linear_probing_and traversal}
\subsection{Monotonic Traversal Hypothesis}
\label{app:monotonic}

The linear probe formulation guarantees monotonic change only for the probe prediction
\(\hat y(z)\). However, the chemically relevant question is stricter: whether moving through the
latent space along the learned direction produces decoded molecules whose measured property
changes monotonically. This is not automatic, because the decoder is nonlinear, many-to-one, and
the target property is evaluated only after decoding the latent point back into a molecular graph.

For a property direction \(w_y\), we define the normalized traversal direction
\[
u_y = \frac{w_y}{\|w_y\|_2}.
\]
Given a seed molecule \(x_0\) with latent code
\[
z_0 = E(x_0),
\]
we construct a one-dimensional latent path
\[
z(t) = z_0 + t u_y,
\]
where \(t\) controls the traversal magnitude and sign. Positive values of \(t\) correspond to moving
in the direction predicted to increase the property, while negative values correspond to moving in
the opposite direction.

This gives a closed-form, non-gradient optimization procedure for the linear surrogate. Indeed,
within a local latent trust region
\[
\mathcal{B}_{\rho}(z_0)
=
\{z : \|z-z_0\|_2 \leq \rho\},
\]
maximizing the probe prediction
\[
\hat y(z)=w_y^\top z+b
\]
has the analytic solution
\[
z^\star
=
z_0+\rho\frac{w_y}{\|w_y\|_2}.
\]

Candidate molecules are obtained by evaluating a one-dimensional path
along \(u_y\), rather than solving a high-dimensional nonlinear optimization problem.

Each point \(z(t)\) is decoded into a SELFIES sequence and converted back into a molecular graph.
The target property is then recomputed from the decoded molecule using the same RDKit descriptor
used to construct the property panel. This gives an observed property trajectory
\[
y(t) = g(D(z(t))),
\]
where \(D\) denotes decoding and \(g\) denotes the graph-based property function. For stochastic
decoding, multiple samples may be generated at each traversal step and summarized by the mean
property among valid decoded molecules,
\[
\bar y(t)
=
\frac{1}{|\mathcal{V}_t|}
\sum_{x \in \mathcal{V}_t}
g(x),
\]
where \(\mathcal{V}_t\) is the set of valid decoded molecules at step \(t\).

The monotonic traversal hypothesis is that, for chemically meaningful global directions, the
decoded property trajectory should vary consistently with the traversal coordinate:
\[
t_1 < t_2
\quad \Longrightarrow \quad
\bar y(t_1) \leq \bar y(t_2)
\]
for increasing directions, up to stochastic decoding noise and local chemical constraints. In
practice, this hypothesis is evaluated using rank correlation between \(t\) and \(\bar y(t)\), the
slope of a linear fit along the traversal path, and the number of monotonicity violations between
successive traversal points.

This test is stricter than held-out probe performance. A high probe \(R^2\) only shows that the
property is linearly decodable from the latent representation. A successful monotonic traversal
shows that the learned direction is also actionable: moving through latent space changes the
decoded molecular distribution in the intended chemical direction. Conversely, if probe
performance is high but traversal is non-monotonic or chemically unstable, then the direction may
encode property information without providing a reliable steering axis.

The same formulation also supports multi-criterion latent exploration. If several properties admit
stable directions, then optimization can be performed in the low-dimensional subspace spanned by
their corresponding vectors. For example, one may move toward increasing drug-likeness while
penalizing synthetic inaccessibility, or search for regions where a target property, synthesizability,
and functional-family retention are jointly satisfied. In this setting, property steering becomes a
geometric operation in latent space: directions define axes of molecular variation, and their
intersections define candidate regions where multiple chemical criteria are simultaneously
favored.

We therefore use monotonic traversal as the operational validation of a global property direction.
A direction is considered chemically meaningful only if it combines three properties: predictive
strength in latent space, robustness to confound controls, and monotonic behavior after decoding.
Additional traversal diagnostics such as validity, uniqueness, similarity to the seed molecule, and
family retention are used to determine whether the property change occurs through coherent local
chemical edits rather than arbitrary jumps in molecular structure.

\subsection{Additional monotonic traversals.}
\begin{figure}[H]
\centering
\includegraphics[width=0.48\textwidth]{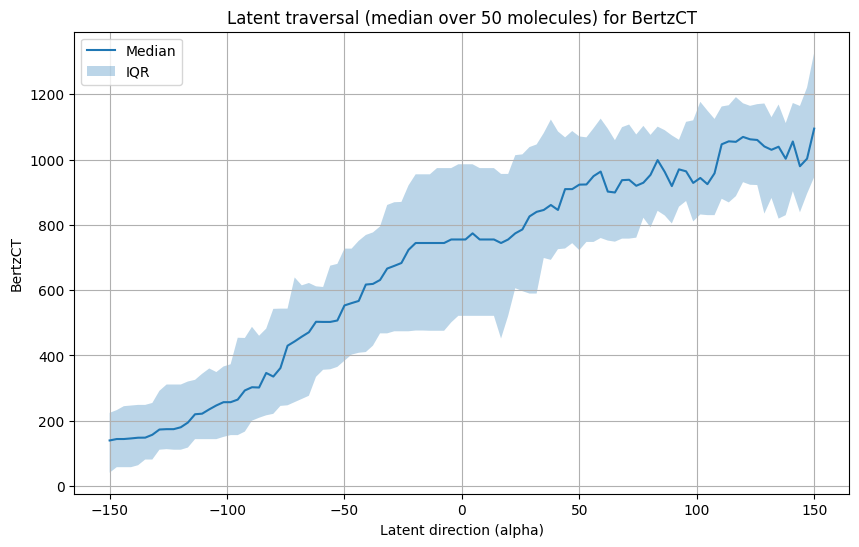}
\includegraphics[width=0.48\textwidth]{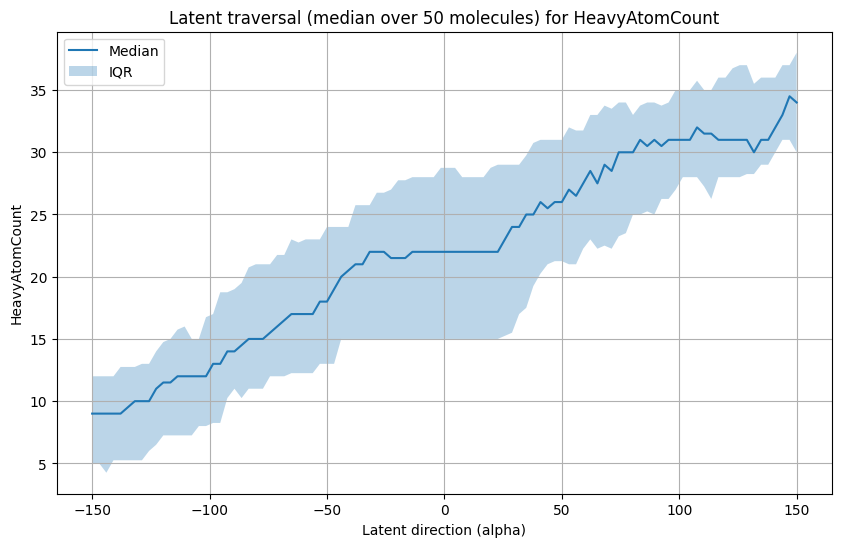}
\caption{Additional monotonic latent traversals from Section~\ref{lineartraversal}. BertzCT and HeavyAtomCount both show broadly increasing median trends across the latent direction, with shaded regions indicating the interquartile range over 50 molecules.}
\label{fig:additional_monotonic_traversals}
\end{figure}

In addition to the properties shown in the main analysis of Section~\ref{lineartraversal}, we observed two further properties with clear monotonic behavior under latent traversal: BertzCT and HeavyAtomCount.
Figure~\ref{fig:additional_monotonic_traversals} highlights that both traversals show a broadly increasing median trend across the latent direction, with the interquartile range indicating that the effect is preserved across the 50 decoded molecules rather than being driven by a small number of outliers. 

\subsection{Inter-Property Correlation and Directional Similarity}
\label{app:interproperty-correlation}

\begin{figure}[H]
    \centering
    \includegraphics[width=\textwidth]{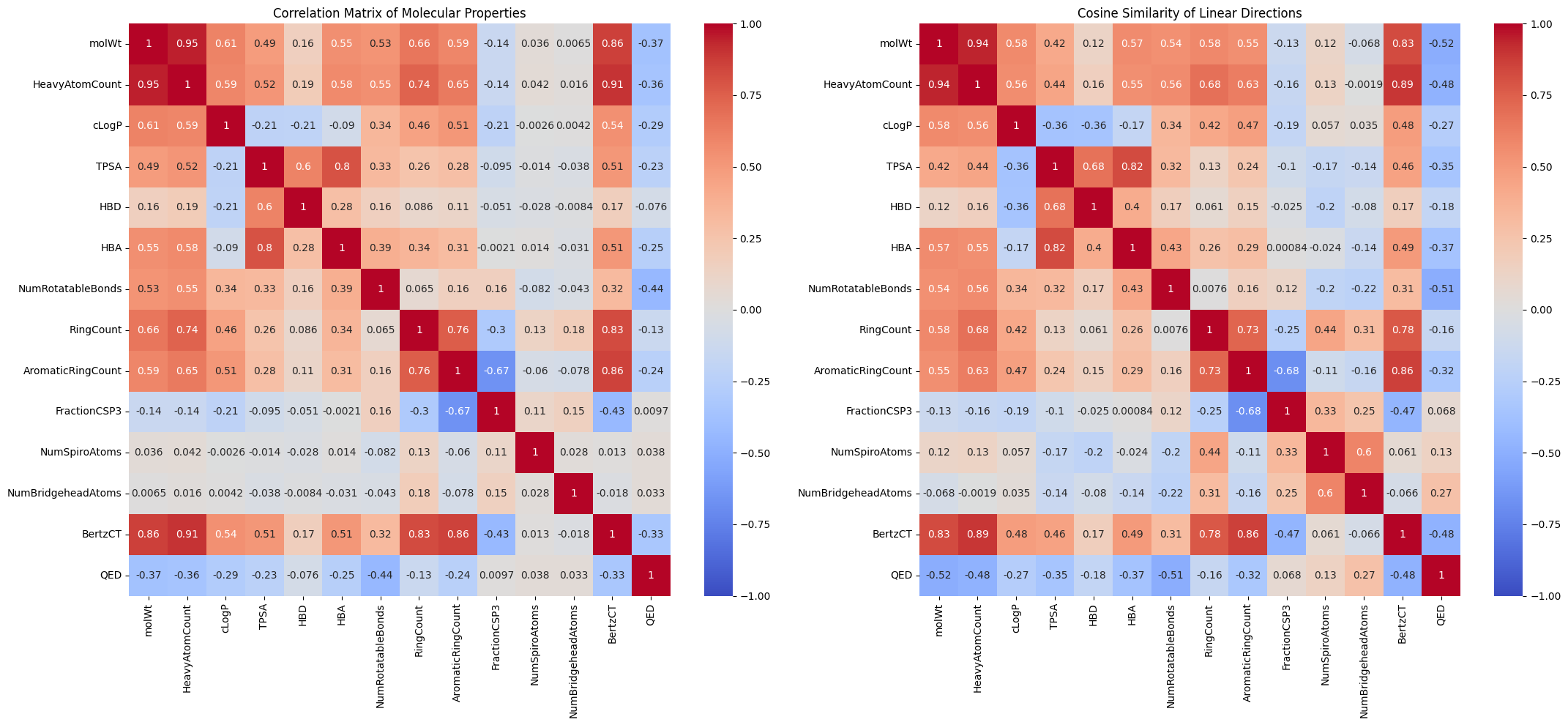}
    \caption{
    Inter-property structure in descriptor space and latent-direction space. 
    Left: empirical Pearson correlation matrix between molecular properties. 
    Right: cosine-similarity matrix between the corresponding linear-probe directions. 
    The broad agreement between the two matrices indicates that the latent geometry preserves meaningful chemical relationships between properties, while discrepancies reveal higher-order organization not fully captured by marginal property correlations.
    }
    \label{fig:interproperty-correlation-directional-similarity}
\end{figure}

To better interpret the traversal results, we compare two complementary objects: the empirical correlation matrix of molecular descriptors and the cosine-similarity matrix between the corresponding linear-probe directions. The first matrix measures how properties co-vary in the dataset, while the second measures whether the latent directions recovered by the probes are geometrically aligned. Both matrices are shown in Figure~\ref{fig:interproperty-correlation-directional-similarity}.

Figure~\ref{fig:interproperty-correlation-directional-similarity} shows that several chemically expected descriptor blocks are preserved in the latent geometry. Size-related descriptors are strongly coupled: \texttt{molWt} and \texttt{HeavyAtomCount} have empirical correlation $0.95$ and directional cosine similarity $0.94$, while \texttt{HeavyAtomCount} and \texttt{BertzCT} have correlation $0.91$ and directional similarity $0.89$. Polar descriptors also form a coherent block, with \texttt{TPSA} and \texttt{HBA} showing correlation $0.80$ and directional similarity $0.82$. Similarly, ring and aromatic descriptors are strongly aligned: \texttt{RingCount}, \texttt{AromaticRingCount}, and \texttt{BertzCT} are mutually correlated, and their probe directions are also highly similar. In particular, \texttt{AromaticRingCount} and \texttt{BertzCT} show both empirical correlation and directional similarity of approximately $0.86$.

This analysis clarifies the interpretation of the confounded descriptors. \texttt{HeavyAtomCount} and especially \texttt{BertzCT} are indeed strongly associated with size-related quantities, but their signal is not reducible to a trivial sequence-length effect. \texttt{BertzCT} aligns not only with \texttt{molWt} and \texttt{HeavyAtomCount}, but also with ring and aromatic descriptors, with directional similarities of $0.78$ with \texttt{RingCount} and $0.86$ with \texttt{AromaticRingCount}. This is consistent with \texttt{BertzCT} acting as a topological complexity descriptor: it combines information about molecular size, branching, ring structure, and aromatic organization. Therefore, the confounding observed for \texttt{BertzCT} should not be interpreted as a simple token-count artifact, but as a structured relationship between sequence-level statistics and chemically meaningful molecular topology.

The behavior of \texttt{FractionCSP3} further supports this interpretation. It is negatively associated with aromaticity and topological complexity, with correlations of $-0.67$ with \texttt{AromaticRingCount} and $-0.43$ with \texttt{BertzCT}. The corresponding latent directions show nearly identical antagonism, with cosine similarities of $-0.68$ and $-0.47$, respectively. Thus, the learned geometry captures the expected opposition between saturated three-dimensional character and aromatic or topologically complex structure.

Some topological descriptors also show stronger directional alignment than raw descriptor correlation. For example, \texttt{NumSpiroAtoms} and \texttt{NumBridgeheadAtoms} have weak empirical correlation ($0.028$), but their latent directions have cosine similarity $0.60$. Likewise, \texttt{RingCount} is only weakly correlated with \texttt{NumSpiroAtoms} ($0.13$), while their latent directions are more clearly aligned ($0.44$). This suggests that the model groups rare ring-topology features in latent space even when they are not strongly correlated at the marginal descriptor level.

Overall, Figure~\ref{fig:interproperty-correlation-directional-similarity} shows that the learned directions are chemically structured rather than independent one-property artifacts. The latent geometry broadly mirrors known descriptor relationships while also revealing higher-order organization among topological features. This supports the interpretation that the successful traversals of \texttt{HeavyAtomCount} and \texttt{BertzCT}, despite their confound sensitivity, arise from a richer latent encoding of molecular topology and complexity rather than from a purely superficial encoding of SELFIES length or token frequency.

\subsection{Ridge vs Linear Regression Ablation Study}
\label{app:ridge-vs-linear}

The main linear-probe study uses ordinary linear regression rather than ridge regression because the fitted probe vector is not used only as a predictive readout. In our setting, the learned weight vector $w$ is also reused as an actionable latent direction for traversal and decoded property steering. This makes the geometry of $w$ part of the method itself.

Ridge regression was initially considered because it often gives strong and stable probe performance in high-dimensional latent spaces. Given latent vectors \(z_i\) and target property values \(y_i\), ridge regression solves
\[
\min_{w,b} \sum_i (y_i - w^\top z_i - b)^2 + \lambda \|w\|^2,
\]
where \(w\) is the learned linear coefficient vector, \(b\) is the bias term, \(i\) indexes the training examples, and \(\lambda \geq 0\) is the regularization strength. The first term measures the squared prediction error, while the second term penalizes large coefficient norms.

This penalty can improve predictive robustness by discouraging overly large weights. However, the same penalty is not geometrically neutral when \(w\) is interpreted as a steering direction in latent space: it may shrink or rotate the fitted vector relative to the unregularized least-squares solution. For this reason, ridge regression is better interpreted here as a regularized predictive probe, whereas ordinary Linear Regression is the cleaner choice for the main steering analysis.

We therefore distinguish the two probes as answering related but different questions:
\[
\text{Ridge: regularized predictive readability},
\]
\[
\text{Linear Regression: unregularized direction for steering}.
\]
This does not make ridge inappropriate. On the contrary, ridge remains useful when predictive accuracy, stability, or property readout is the primary objective. We include it as a complementary sensitivity analysis to verify that the latent-property conclusions are not specific to the unregularized probe.

\begin{table}[H]
\centering
\caption{Ridge versus ordinary Linear Regression for the autoregressive MultiSlotting model. Both raw and residual $R^2$ are reported. $\Delta$ denotes ridge minus Linear Regression.}
\label{tab:app_ridge_vs_linear_ar}
\resizebox{\textwidth}{!}{
\begin{tabular}{lcccccc}
\toprule
Property 
& Linear raw 
& Ridge raw 
& $\Delta_{\mathrm{raw}}$ 
& Linear resid. 
& Ridge resid. 
& $\Delta_{\mathrm{resid}}$ \\
\midrule
HeavyAtomCount      & 0.9750 & 0.9884 & +0.0134 & 0.6554 & 0.7692 & +0.1138 \\
MolWt               & 0.9480 & 0.9655 & +0.0175 & 0.5918 & 0.7000 & +0.1082 \\
BertzCT             & 0.9512 & 0.9677 & +0.0165 & 0.7009 & 0.7537 & +0.0528 \\
RingCount           & 0.9378 & 0.9567 & +0.0189 & 0.7944 & 0.8506 & +0.0562 \\
AromaticRingCount   & 0.8937 & 0.9090 & +0.0153 & 0.7559 & 0.7767 & +0.0208 \\
cLogP               & 0.7967 & 0.8442 & +0.0475 & 0.6861 & 0.7552 & +0.0691 \\
TPSA                & 0.8059 & 0.8670 & +0.0611 & 0.6863 & 0.7728 & +0.0865 \\
HBA                 & 0.8008 & 0.8389 & +0.0381 & 0.6759 & 0.7138 & +0.0379 \\
FractionCSP3        & 0.8951 & 0.9153 & +0.0202 & 0.8313 & 0.8542 & +0.0229 \\
HBD                 & 0.4997 & 0.5535 & +0.0538 & 0.4624 & 0.5211 & +0.0587 \\
NumRotatableBonds   & 0.6914 & 0.7640 & +0.0726 & 0.4752 & 0.5960 & +0.1208 \\
NumSpiroAtoms       & 0.0878 & 0.1073 & +0.0195 & 0.0728 & 0.0914 & +0.0186 \\
NumBridgeheadAtoms  & 0.1522 & 0.1878 & +0.0356 & 0.1207 & 0.1556 & +0.0349 \\
QED                 & 0.4624 & 0.5121 & +0.0497 & 0.3907 & 0.4424 & +0.0517 \\
\midrule
Mean                & 0.7070 & 0.7412 & +0.0343 & 0.5643 & 0.6252 & +0.0609 \\
\bottomrule
\end{tabular}
}
\end{table}

Table \ref{tab:app_ridge_vs_linear_ar} shows that, as expected, ridge generally improves predictive $R^2$, with an average gain of $0.0343$ in the raw setting and $0.0609$ after residualization. The gain is most visible for residualized size- or topology-related descriptors such as HeavyAtomCount, MolWt, and NumRotatableBonds, where regularization stabilizes the readout. Importantly, however, the global interpretation is unchanged. FractionCSP3, TPSA, cLogP, and HBA remain strongly predictable under both probes, while NumSpiroAtoms and NumBridgeheadAtoms remain weak under linear readout. Thus, ridge improves predictive readability but does not alter the qualitative distinction between globally steerable descriptors and descriptors that likely require nonlinear or local structure.

\begin{table}[H]
\centering
\caption{Ridge-probe comparison across the three latent models. Raw and residual $R^2$ are reported for the same descriptor panel used in the previous main-text analysis. Best values within each target setting are bolded.}
\label{tab:app_ridge_three_model_comparison}
\resizebox{\textwidth}{!}{
\begin{tabular}{lcccccc}
\toprule
Property 
& Linear Attn. raw 
& Simple Attn. raw 
& AR raw 
& Linear Attn. resid. 
& Simple Attn. resid. 
& AR resid. \\
\midrule
HeavyAtomCount      & 0.9759 & 0.9622 & \textbf{0.9884} & 0.4857 & 0.3787 & \textbf{0.7692} \\
MolWt               & 0.9502 & 0.9190 & \textbf{0.9655} & 0.5808 & 0.3170 & \textbf{0.7000} \\
BertzCT             & 0.9494 & 0.9303 & \textbf{0.9677} & 0.6564 & 0.5646 & \textbf{0.7537} \\
RingCount           & 0.8918 & 0.8495 & \textbf{0.9567} & 0.6013 & 0.5154 & \textbf{0.8506} \\
AromaticRingCount   & 0.8855 & 0.8503 & \textbf{0.9090} & 0.7210 & 0.6577 & \textbf{0.7767} \\
cLogP               & 0.7699 & 0.6824 & \textbf{0.8442} & 0.6572 & 0.5791 & \textbf{0.7552} \\
TPSA                & 0.7765 & 0.7008 & \textbf{0.8670} & 0.6307 & 0.5251 & \textbf{0.7728} \\
HBA                 & 0.7998 & 0.7602 & \textbf{0.8389} & 0.6360 & 0.5427 & \textbf{0.7138} \\
FractionCSP3        & 0.8587 & 0.7862 & \textbf{0.9153} & 0.7843 & 0.7161 & \textbf{0.8542} \\
HBD                 & 0.3518 & 0.2627 & \textbf{0.5535} & 0.3100 & 0.2226 & \textbf{0.5211} \\
NumSpiroAtoms       & 0.0666 & 0.0513 & \textbf{0.1073} & 0.0504 & 0.0358 & \textbf{0.0914} \\
NumBridgeheadAtoms  & 0.1070 & 0.0816 & \textbf{0.1878} & 0.0781 & 0.0490 & \textbf{0.1556} \\
QED                 & 0.4308 & 0.3455 & \textbf{0.5121} & 0.3652 & 0.2789 & \textbf{0.4424} \\
\midrule
Mean                & 0.6780 & 0.6294 & \textbf{0.7395} & 0.5044 & 0.4141 & \textbf{0.6274} \\
\bottomrule
\end{tabular}
}
\end{table}

Moreover, the ridge comparison in Table \ref{tab:app_ridge_three_model_comparison} confirms that the autoregressive MultiSlotting model remains the strongest representation even under a regularized predictive probe. It obtains the best raw and residual $R^2$ across the descriptor panel, including both confound-sensitive descriptors such as HeavyAtomCount, MolWt, and BertzCT, and chemically meaningful residual descriptors such as FractionCSP3, TPSA, cLogP, and HBA. This supports the same qualitative conclusion as the main Linear Regression analysis: the autoregressive latent space is more readable and better organized than the ablation models.

Overall, the ridge ablation reinforces the robustness of the latent interpretability results while clarifying the methodological role of the main probe. Ridge is useful as a prediction-oriented sensitivity analysis, but the main steerability claims are based on ordinary Linear Regression because the learned direction is used directly for latent traversal and decoded molecular steering.

\subsection{Simple and linear attention models and their latent traversals}
\label{app:linearattention}

For completeness, we also evaluated latent traversals for the linear-attention and simple-attention baselines. These experiments are useful because they test a stricter condition than probe accuracy: not only whether a property is decodable from the latent representation, but whether the corresponding probe direction behaves like a usable control axis after decoding back to molecules.

\begin{figure}[t]
\centering
\includegraphics[width=0.5\textwidth]{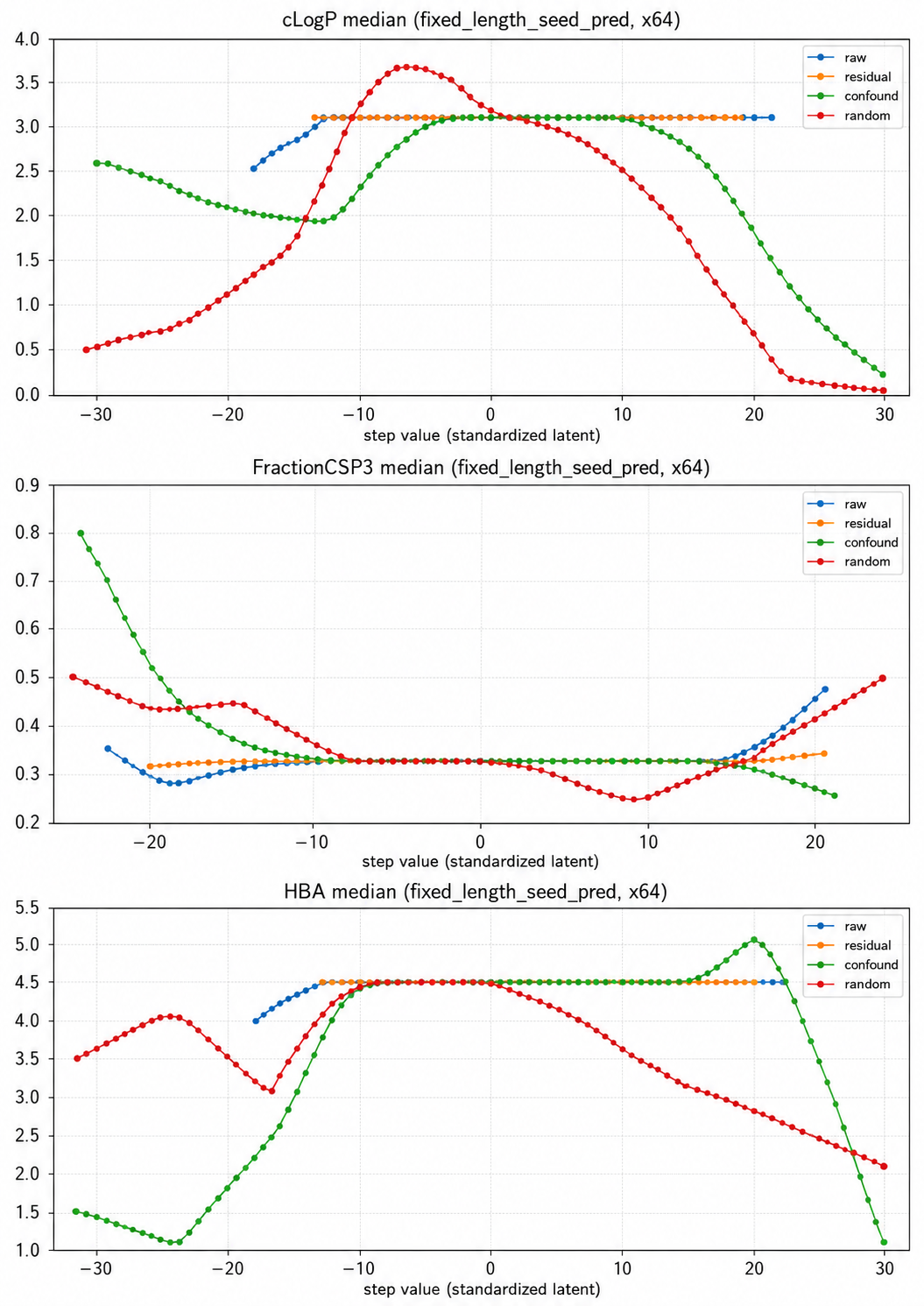}
\caption{Raw, confound, residual, and random-axis latent traversals for the linear-attention model. The decoded properties are mostly flat near the seed and exhibit abrupt, saturated, or non-monotonic changes rather than stable directional control.}
\label{fig:linear_attention_traversals}
\end{figure}

Figure~\ref{fig:linear_attention_traversals} shows representative traversals for the linear-attention model on cLogP, F.CSP3, and HBA. The dominant pattern is not a smooth property gradient, but a combination of plateaus, cliffs, and saturation. Around the seed, decoded molecules often remain effectively unchanged over a wide range of step sizes, suggesting that movement along the latent direction is absorbed by the decoder without producing chemically distinct outputs. When the property does change, the transition is frequently abrupt or non-monotonic, and the targeted raw direction is not cleanly separated from residual, confound, or random directions. In other words, the property signal may be present in the latent state, but it is not organized as a stable, axis-aligned direction that supports controlled molecular editing.

\begin{figure}[t]
\centering
\includegraphics[width=0.5\textwidth]{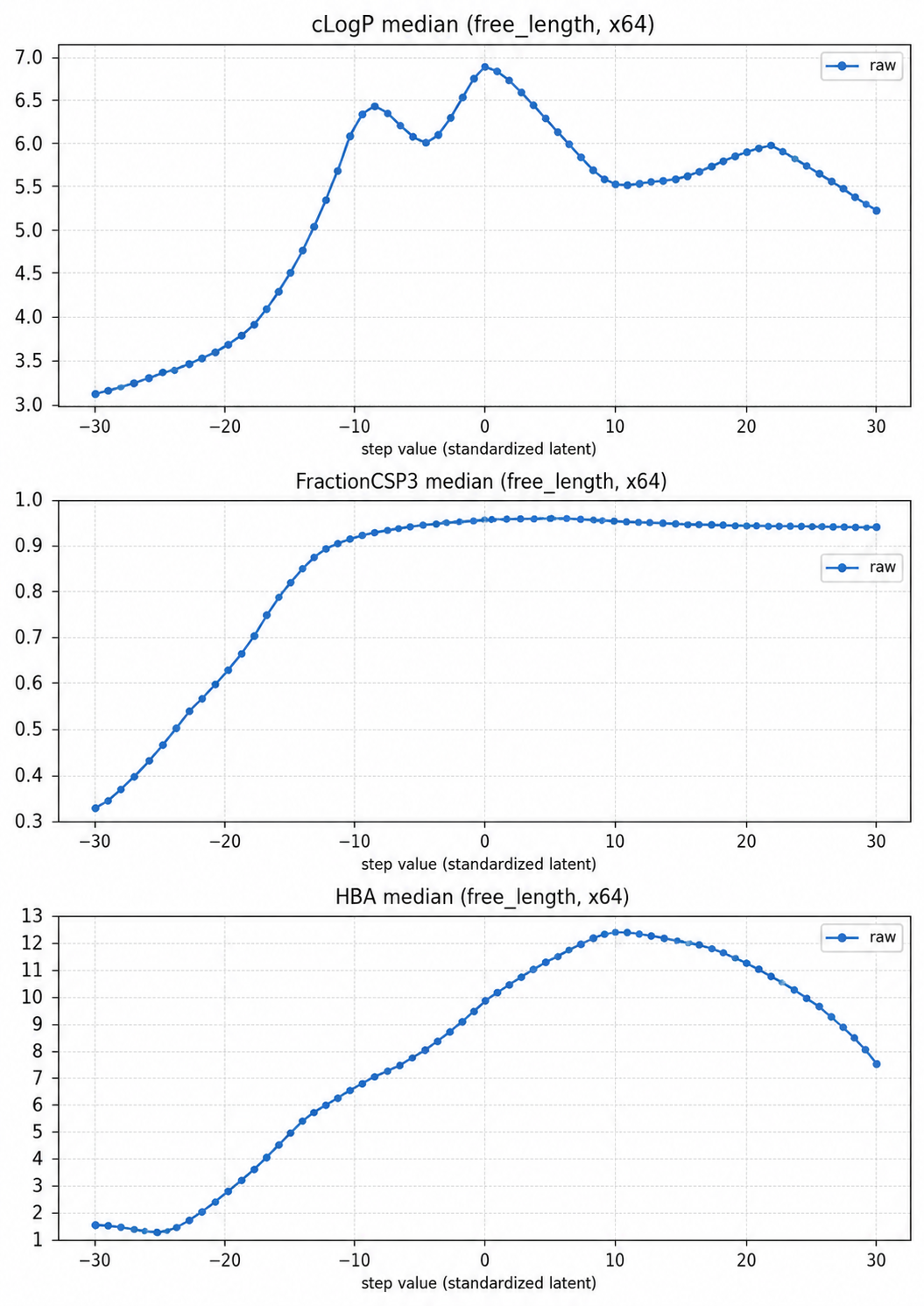}
\caption{Raw-axis latent traversals for the simple-attention model. The simple-attention baseline produces clearer property responses than the linear-attention model, but the trajectories show saturation, non-monotonicity, and dependence on traversal scale.}
\label{fig:simple_attention_traversals}
\end{figure}

Figure~\ref{fig:simple_attention_traversals} shows the corresponding raw-axis traversals for the simple-attention model. This baseline is more responsive: cLogP, F.CSP3, and HBA all show clear property movement along the raw probe direction. However, the response is still not consistently traversal-ready. F.CSP3 rapidly increases and then saturates, indicating that the direction captures a coarse transition rather than a continuously tunable property. HBA shows a large increase over part of the traversal but later bends back downward, while cLogP exhibits pronounced scale dependence and non-monotonic behavior. Thus, the simple-attention latent space contains stronger chemically meaningful signal than the linear-attention model, but that signal is only locally useful and does not define a globally reliable control coordinate.

Together, these baselines highlight the distinction between \emph{latent decodability} and \emph{latent controllability}. A probe can recover property information from a representation even when moving along its direction does not induce a smooth or monotonic change in decoded molecules. The attention baselines therefore encode chemically relevant information, but their latent geometry is less aligned with direct molecular control than the autoregressive model. In the autoregressive setting, probe directions behave more like usable editing axes; in the attention baselines, they behave more like correlational readout directions whose effect on decoded chemistry is discontinuous, saturated, or strongly scale-dependent.

\section{Non-Linear probing}
\subsection{Gradient-Based Local Directions}
\label{app:nonlinear gradient}

Unlike the linear probe, the MLP does not define a single global direction. Its local direction at a latent point is given by the gradient of the learned predictor,
\[
g_y(\tilde z)
=
\nabla_{\tilde z} f_{\phi,y}(\tilde z),
\]
where \(\tilde z\) denotes the standardized latent vector, \(f_{\phi,y}\) is the MLP predictor for property \(y\) with learned parameters \(\phi\), and \(g_y(\tilde z)\) is the gradient of that predictor with respect to \(\tilde z\).

The corresponding normalized local steering direction is
\[
u_y(\tilde z)
=
\frac{
g_y(\tilde z)
}{
\|g_y(\tilde z)\|_2 + \varepsilon
},
\]
where \(u_y(\tilde z)\) is the unit-normalized local direction, \(\|\cdot\|_2\) denotes the Euclidean norm, and \(\varepsilon>0\) is a small constant used to avoid division by zero.

A nonlinear traversal is therefore defined recursively as
\[
\tilde z_{k+1}
=
\tilde z_k
+
\eta
u_y(\tilde z_k),
\]
where \(k\) indexes the traversal step and \(\eta>0\) is the step size. Traversal in the decreasing direction is obtained by replacing \(+\eta\) with \(-\eta\). Since the MLP is trained in standardized latent coordinates, the traversed point is mapped back to the decoder space by
\[
z_k
=
\mu_Z + s_Z \odot \tilde z_k,
\]
where \(z_k\) is the corresponding latent vector in the decoder coordinate system, \(\mu_Z\) and \(s_Z\) are the empirical mean and standard deviation of the latent vectors used for standardization, and \(\odot\) denotes elementwise multiplication.

This makes nonlinear steering a path through a learned vector field rather than a straight-line displacement. The linear probe gives
\[
\nabla_z(w^\top z+b)=w,
\]
where \(w\) and \(b\) are the linear probe coefficient vector and bias term. Thus, every molecule shares the same direction. The MLP gives
\[
\nabla_{\tilde z} f_{\phi}(\tilde z),
\]
where \(f_{\phi}\) denotes the learned nonlinear predictor, so the direction depends on the current molecule. Local steering is therefore most appropriate for properties that are predictable from \(z\) but not organized around a stable global axis.

As with linear traversal, decoded molecules along the path are evaluated using the external RDKit property function,
\[
y_k
=
g(D(z_k)),
\]
where \(D\) is the trained decoder, \(D(z_k)\) is the molecule decoded from the latent vector \(z_k\), \(g\) is the external RDKit property evaluator, and \(y_k\) is the resulting property value at traversal step \(k\). The MLP gradient is therefore treated as a steering hypothesis whose chemical validity must be verified after decoding.

\subsubsection{Interpreting Delta-$R^2$}
\label{app:delta_r2}

To distinguish globally linear organization from merely decodable information, we define
\[
\Delta R^2 = R^2_{\mathrm{MLP}} - R^2_{\mathrm{linear}}.
\]
Small values indicate that a linear probe already captures most of the available property signal, whereas large values indicate additional nonlinear structure accessible only to the MLP.

Table~\ref{tab:delta_r2_expanded_properties} reports the Delta-$R^2$ analysis on the expanded property set. This diagnostic supports the traversal results. The descriptors that showed monotonic behavior under linear latent traversal; HeavyAtomCount, BertzCT, FractionCSP3, TPSA, cLogP, and HBA, also have comparatively low $\Delta R^2$. In the autoregressive model, their mean $\Delta R^2$ is $0.061$ in the raw setting and $0.152$ after residualization. By contrast, the nonlinear descriptor group \{HBD, NumRotatableBonds, NumSpiroAtoms, NumBridgeheadAtoms, QED\} has much larger mean gaps, $0.360$ raw and $0.396$ residual. The same trend appears in the simple-attention model, but with larger gaps for the monotonic descriptors, increasing to $0.134$ raw and $0.305$ residual, and a much larger gap for the nonlinear descriptor ($0.410$ raw and $0.453$ residual).

Thus, both models encode chemically meaningful descriptor information, but the autoregressive latent space organizes the traversal-relevant descriptors in a more globally linear way. This explains why these properties are more stable under linear latent traversal. AromaticRingCount is excluded from the clean monotonic group because, although its $\Delta R^2$ is small, the confound analysis showed substantial correlation with ring-token features. SA is also not treated as a clean traversal target because its traversal behavior is unstable under abrupt structural changes.

\begin{table}[H]
\centering
\caption{Delta-$R^2$ analysis on the expanded property set. Low $\Delta R^2$ indicates that the linear probe already captures most of the available signal; high $\Delta R^2$ indicates substantial nonlinear structure.}
\label{tab:delta_r2_expanded_properties}
\resizebox{\textwidth}{!}{
\begin{tabular}{llcccc}
\toprule
Property & Regime / interpretation & AR $\Delta R^2_{\text{raw}}$ & AR $\Delta R^2_{\text{resid}}$ & Simple $\Delta R^2_{\text{raw}}$ & Simple $\Delta R^2_{\text{resid}}$ \\
\midrule

F.CSP3    & monotonic linear regime & 0.067 & 0.111 & 0.175 & 0.220 \\
TPSA      & monotonic linear regime & 0.078 & 0.143 & 0.206 & 0.319 \\
cLogP     & monotonic linear regime & 0.089 & 0.145 & 0.207 & 0.260 \\
HBA       & monotonic linear regime & 0.104 & 0.180 & 0.146 & 0.294 \\
HAC       & monotonic linear regime (size confounded) & 0.007 & 0.163 & 0.028 & 0.449 \\
BertzCT   & monotonic linear regime (size confounded)& 0.021 & 0.169 & 0.044 & 0.289 \\
MolWt     & confounded regime (size confounded) & 0.024 & 0.235 & 0.062 & 0.523 \\
Ring      & confounded regime (ring confounded)& 0.029 & 0.107 & 0.104 & 0.340 \\
A.Ring    & confounded regime (ring confounded)& 0.047 & 0.126 & 0.083 & 0.205 \\
HBD       & nonlinear structure & 0.237 & 0.256 & 0.385 & 0.406 \\
Spiro     & nonlinear structure & 0.517 & 0.513 & 0.385 & 0.388 \\
Bridge    & nonlinear structure & 0.592 & 0.610 & 0.603 & 0.612 \\
QED       & nonlinear structure & 0.307 & 0.345 & 0.339 & 0.359 \\
NumRotBond & nonlinear structure & 0.145 & 0.253 & 0.337 & 0.502 \\
SA        & unstable traversal target & 0.160 & 0.180 & 0.254 & 0.286 \\
\midrule
Mean over monotonic descriptors 
& HAC, BertzCT, F.CSP3, TPSA, cLogP, HBA & 0.061 & 0.152 & 0.134 & 0.305 \\
Mean over nonlinear descriptors
& HBD, Spiro, Bridge, QED, NumRotBond & 0.360 & 0.396 & 0.410 & 0.453 \\
\bottomrule
\end{tabular}
}
\end{table}

\newpage
\section*{NeurIPS Paper Checklist}

\begin{enumerate}

\item {\bf Claims}
    \item[] Question: Do the main claims made in the abstract and introduction accurately reflect the paper's contributions and scope?
    \item[] Answer: \answerYes{} 
    \item[] Justification: We claim that an unsupervised SELFIES Transformer-VAE can learn a latent space in which several RDKit descriptors admit approximately linear, decoded-molecule steering directions. We support this through confound-aware probing and traversal experiments.

\item {\bf Limitations}
    \item[] Question: Does the paper discuss the limitations of the work performed by the authors?
    \item[] Answer: \answerYes{} 
    \item[] Justification: We provide a short discussion about the limitations in the "Conclusion, limitations and future directions" section
    
\item {\bf Theory assumptions and proofs}
    \item[] Question: For each theoretical result, does the paper provide the full set of assumptions and a complete (and correct) proof?
    \item[] Answer: \answerYes{} 
    \item[] Justification: The paper makes a limited theoretical claim rather than introducing a new general theory: for a linear property probe, the probe weight vector defines the first-order direction of monotonic change in the predicted property. The assumptions and derivations for this statement, as well as for the residualization procedure used to remove confound-predictable components, are given in the theory section and Appendix. The theory is then validated experimentally through latent traversal results.

    \item {\bf Experimental result reproducibility}
    \item[] Question: Does the paper fully disclose all the information needed to reproduce the main experimental results of the paper to the extent that it affects the main claims and/or conclusions of the paper (regardless of whether the code and data are provided or not)?
    \item[] Answer: \answerYes{} 
    \item[] Justification: The sections "Experimental setup" in the main paper, and "Experimental Setup and Reproducibility" in the Appendix, provide all necessary information to reproduce the experiments

\item {\bf Open access to data and code}
    \item[] Question: Does the paper provide open access to the data and code, with sufficient instructions to faithfully reproduce the main experimental results, as described in supplemental material?
    \item[] Answer: \answerYes{} 
    \item[] Justification: A well documented, anonymized github repo containing both the code and the processed data is shared in subsection "Experimental Setup and Reproducibility" in the Appendix.

\item {\bf Experimental setting/details}
    \item[] Question: Does the paper specify all the training and test details (e.g., data splits, hyperparameters, how they were chosen, type of optimizer) necessary to understand the results?
    \item[] Answer: \answerYes{}{} 
    \item[] Justification: All relevant info is shared  in subsection "Experimental setup" of the main paper and in subsection "Experimental Setup and Reproducibility" in the Appendix.
    
\item {\bf Experiment statistical significance}
    \item[] Question: Does the paper report error bars suitably and correctly defined or other appropriate information about the statistical significance of the experiments?
    \item[] Answer: \answerYes{} 
    \item[] Justification: The paper reports uncertainty and robustness information beyond single-run results. Traversal analyses are evaluated over multiple molecule seeds rather than a single example, and aggregate statistics are reported for traversal behavior, family retention, and latent generation. In addition, bootstrap estimates, permutation controls, and random-direction controls are used where appropriate to quantify stability and distinguish meaningful latent directions from chance effects.

\item {\bf Experiments compute resources}
    \item[] Question: For each experiment, does the paper provide sufficient information on the computer resources (type of compute workers, memory, time of execution) needed to reproduce the experiments?
    \item[] Answer: \answerYes{}{} 
    \item[] Justification: This is shared in subsection "Experimental Setup and Reproducibility" in the Appendix.
    
\item {\bf Code of ethics}
    \item[] Question: Does the research conducted in the paper conform, in every respect, with the NeurIPS Code of Ethics \url{https://neurips.cc/public/EthicsGuidelines}?
    \item[] Answer: \answerYes{} 
    \item[] Justification: The study uses molecular-structure data and computed RDKit descriptors only, and does not involve human subjects, private data, surveillance data, or protected attributes. We credit and license the dataset and released artifacts, provide reproducibility details, and do not make claims involving deployment, clinical use, or validated biological activity, in line with the NeurIPS Code of Ethics’ requirements on data use, reproducibility, and potential harms.

\item {\bf Broader impacts}
    \item[] Question: Does the paper discuss both potential positive societal impacts and negative societal impacts of the work performed?
    \item[] Answer: \answerNo{} 
    \item[] Justification: The paper does not explicitly discuss societal impacts because the contribution is limited to representation analysis of molecular VAE latent spaces. The work studies whether known molecular descriptors admit stable linear traversal directions; it does not present a deployable molecule-design pipeline, optimize biological activity, or validate generated compounds for synthesis or use. As a result, the immediate societal impact, especially negative impact, is limited. Potential positive implications for future drug or materials discovery are indirect rather than a central claim of the paper.

\item {\bf Safeguards}
    \item[] Question: Does the paper describe safeguards that have been put in place for responsible release of data or models that have a high risk for misuse (e.g., pre-trained language models, image generators, or scraped datasets)?
    \item[] Answer: \answerNA{} 
    \item[] Justification: The released assets are intended for reproducible representation analysis and descriptor-level latent steering, not for deployment, synthesis planning, biological activity optimization, toxicity optimization, or validated molecule design. The dataset contains molecular structures rather than scraped personal or sensitive data, and we do not identify the release as high-risk or dual-use in the sense requiring controlled access safeguards.

\item {\bf Licenses for existing assets}
    \item[] Question: Are the creators or original owners of assets (e.g., code, data, models), used in the paper, properly credited and are the license and terms of use explicitly mentioned and properly respected?
    \item[] Answer: \answerYes{} 
    \item[] Justification: The dataset is openly available under CDLA-Sharing-1.0, a data-sharing license that permits use, modification, and redistribution under reciprocal sharing terms, and it has been properly credited. The newly released code, checkpoints, and processed artifacts are documented in the anonymized repository and released for public use.

\item {\bf New assets}
    \item[] Question: Are new assets introduced in the paper well documented and is the documentation provided alongside the assets?
    \item[] Answer: \answerYes{} 
    \item[] Justification: Everything is well documented in the "Experimental Setup and Reproducibility" subsection of the Appendix and in the anonymized github repo.

\item {\bf Crowdsourcing and research with human subjects}
    \item[] Question: For crowdsourcing experiments and research with human subjects, does the paper include the full text of instructions given to participants and screenshots, if applicable, as well as details about compensation (if any)? 
    \item[] Answer: \answerNA{} 
    \item[] Justification: The paper does not involve crowdsourcing nor research with human subjects.

\item {\bf Institutional review board (IRB) approvals or equivalent for research with human subjects}
    \item[] Question: Does the paper describe potential risks incurred by study participants, whether such risks were disclosed to the subjects, and whether Institutional Review Board (IRB) approvals (or an equivalent approval/review based on the requirements of your country or institution) were obtained?
    \item[] Answer: \answerNA{} 
    \item[] Justification: The paper does not involve crowdsourcing nor research with human subjects.

\item {\bf Declaration of LLM usage}
    \item[] Question: Does the paper describe the usage of LLMs if it is an important, original, or non-standard component of the core methods in this research? Note that if the LLM is used only for writing, editing, or formatting purposes and does \emph{not} impact the core methodology, scientific rigor, or originality of the research, declaration is not required.
    \item[] Answer: \answerNA{} 
    \item[] Justification: The core method development in this research does not involve LLMs as any important, original, or non-standard components.

\end{enumerate}

\end{document}